\documentclass[final]{cvpr}

\usepackage{times}
\usepackage{epsfig}
\usepackage{graphicx}
\usepackage{amsmath}
\usepackage{amssymb}

\usepackage{enumitem}

\usepackage{animate}
\usepackage{subfig}
\usepackage{array}
\usepackage{multirow}
\usepackage{mathabx}
\usepackage{microtype}
\usepackage{enumitem}
\usepackage{xspace}
\usepackage{xcolor}
\usepackage{color,soul}
\usepackage{multicol}
\usepackage{colortbl}
\usepackage{tabularx,booktabs}
\usepackage{cancel}
\usepackage{booktabs}
\usepackage{caption}
\usepackage[nocompress]{cite}

\usepackage{float}

\let\oldFootnote\footnote
\newcommand\nextToken\relax
\renewcommand\footnote[1]{%
    \oldFootnote{#1}\futurelet\nextToken\isFootnote}
\newcommand\isFootnote{%
    \ifx\footnote\nextToken\textsuperscript{,}\fi}

\definecolor{overview_orange}{HTML}{FFA726}
\definecolor{overview_green}{HTML}{9CCC65}
\definecolor{overview_purple}{HTML}{D877E9}
\definecolor{overview_red}{HTML}{EF5350}
\definecolor{overview_gray}{HTML}{BEBEBE}

\newcolumntype{Y}{>{\centering\arraybackslash}X}

\makeatletter
\let\UrlSpecialsOld\UrlSpecials
\def\UrlSpecials{\UrlSpecialsOld\do\/{\Url@slash}\do\_{\Url@underscore}}%
\def\Url@slash{\@ifnextchar/{\kern-.11em\mathchar47\kern-.2em}%
    {\kern-.0em\mathchar47\kern-.08em\penalty\UrlBigBreakPenalty}}
\def\Url@underscore{\nfss@text{\leavevmode \kern.06em\vbox{\hrule\@width.3em}}}
\makeatother

\usepackage{array}
\newcommand{\PreserveBackslash}[1]{\let\temp=\\#1\let\\=\temp}
\newcolumntype{C}[1]{>{\PreserveBackslash\centering}p{#1}}
\newcolumntype{R}[1]{>{\PreserveBackslash\raggedleft}p{#1}}
\newcolumntype{L}[1]{>{\PreserveBackslash\raggedright}p{#1}}

\def\arxiv{for arxiv submission}

\ifdefined\arxiv
\newcommand{\acrobat}{\emph{Click the image to play the video in a browser.}}
\newcommand{\supplementary}[1]{Appendix #1}

\else
\newcommand{\acrobat}{\emph{The figure is best viewed with Acrobat Reader. Click each image to play the video clip.}}
\newcommand{\supplementary}[1]{Appendix #1}
\def\embedVideo{embed videos in pdf}

\fi

\usepackage[pagebackref=true,breaklinks=true,letterpaper=true,colorlinks,bookmarks=false]{hyperref}

\def\cvprPaperID{0517} 
\def\confYear{CVPR 2021}

\begin{document}

\title{One-Shot Free-View Neural Talking-Head Synthesis for Video Conferencing}

\author{Ting-Chun Wang \quad  Arun Mallya \quad  Ming-Yu Liu \\ \\ NVIDIA Corporation}

\twocolumn[{%
\renewcommand\twocolumn[1][]{#1}%
\maketitle
\begin{center}

\vspace{-.2in}
\begin{center}
    \begin{minipage}[c]{\textwidth}
    \begin{tabular}{ ccccc }
     \multicolumn{3}{c}{
     \ifdefined\embedVideo
     \animategraphics[width=0.59\textwidth]{25}{figures/teaser/compression/}{00000}{00089}
     \else
     \ifdefined\embedShortVideo
     \animategraphics[width=0.59\textwidth]{25}{figures/teaser/compression/}{00015}{00089}
     \else
     \href{https://nvlabs.github.io/face-vid2vid/web_gifs/teaser_compression.gif}{\includegraphics[width=.59\textwidth]{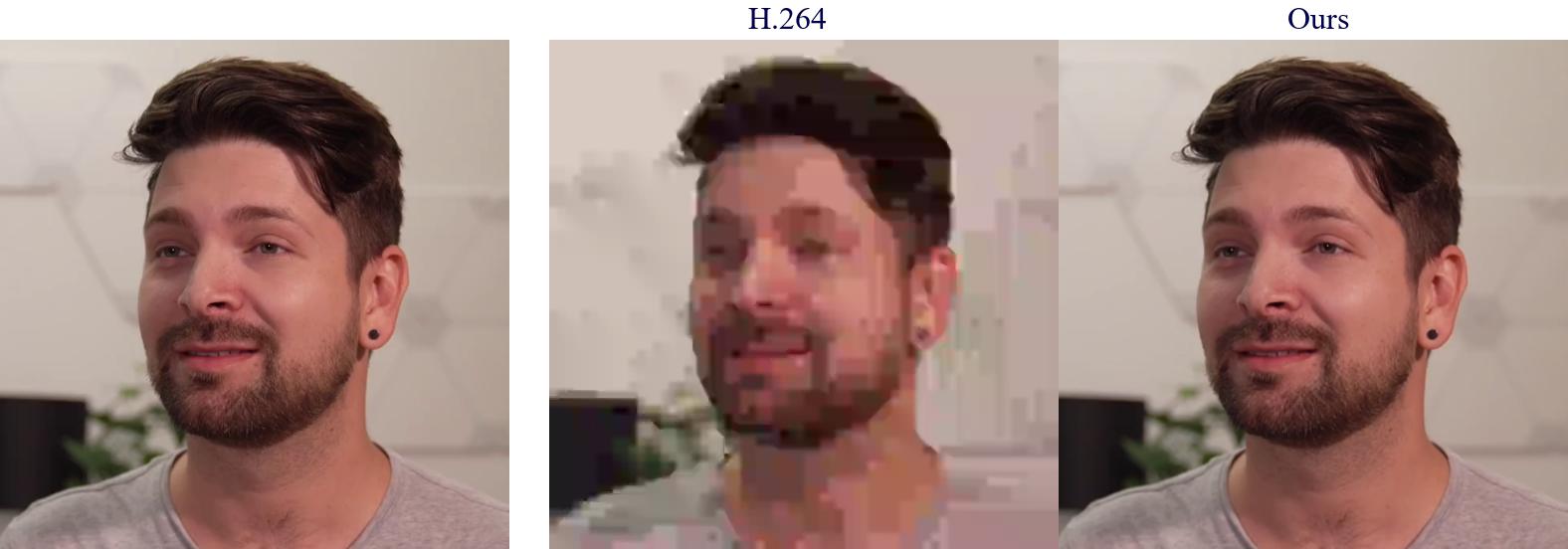}}
     \fi} &
     \multicolumn{2}{c}{
     \ifdefined\embedVideo
     \animategraphics[width=0.38\textwidth]{25}{figures/teaser/novel_view/}{00000}{00089}
     \else
     \ifdefined\embedShortVideo
     \animategraphics[width=0.38\textwidth]{25}{figures/teaser/novel_view/}{00015}{00089}
     \else
     \href{https://nvlabs.github.io/face-vid2vid/web_gifs/teaser_novelview.gif}{\includegraphics[width=.38\textwidth]{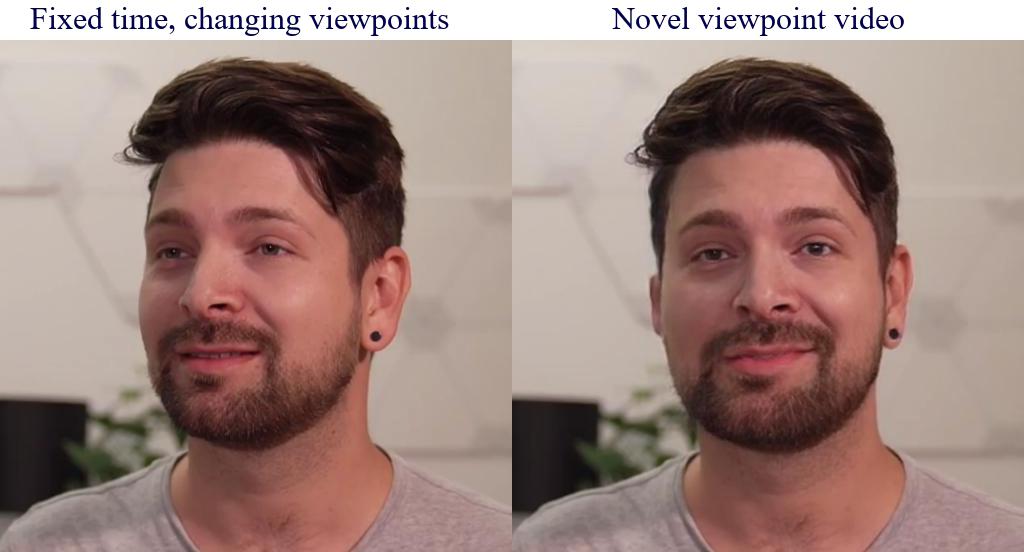}}
     \fi} \\
     \hspace{4mm} (a) Original video &
     \multicolumn{2}{c}{\hspace{1mm} (b) Compressed videos at the same bit-rate} &
     \multicolumn{2}{c}{(c) Our re-rendered novel-view results}
    \end{tabular}
    \end{minipage}
    \vspace{-.1in}
    \captionof{figure}{Our method can re-create a talking-head video using only a single source image (\eg, the first frame) and a sequence of unsupervisedly-learned 3D keypoints, representing motions in the video. Our novel keypoint representation provides a compact representation of the video that is 10$\times$ more efficient than the H.264 baseline can provide. A novel 3D keypoint decomposition scheme allows re-rendering the talking-head video under different poses, simulating often missed face-to-face video conferencing experiences. \emph{\textbf{\acrobat}}}
	\label{figure:teaser}
\end{center}

\end{center}%
}]
\maketitle


\begin{abstract}
\vspace{-.1in}
We propose a neural talking-head video synthesis model and demonstrate its application to video conferencing. Our model learns to synthesize a talking-head video using a source image containing the target person's appearance and a driving video that dictates the motion in the output. Our motion is encoded based on a novel keypoint representation, where the identity-specific and motion-related information is decomposed unsupervisedly. Extensive experimental validation shows that our model outperforms competing methods on benchmark datasets. Moreover, our compact keypoint representation enables a video conferencing system that achieves the same visual quality as the commercial H.264 standard while only using one-tenth of the bandwidth. Besides, we show our keypoint representation allows the user to rotate the head during synthesis, which is useful for simulating face-to-face video conferencing experiences. Our project page can be found at \url{https://nvlabs.github.io/face-vid2vid}.
\end{abstract}

\vspace{-.2in}
\section{Introduction}

We study the task of generating a realistic talking-head video of a person using one source image of that person and a driving video, possibly derived from another person. The source image encodes the target person's appearance, and the driving video dictates motions in the output video.

We propose a \textit{pure} neural rendering approach, where we render a talking-head video using a deep network in the one-shot setting without using a graphics model of the 3D human head. Compared to 3D graphics-based models, 2D-based methods enjoy several advantages. First, it avoids 3D model acquisition, which is often laborious and expensive. Second, 2D-based methods can better handle the synthesis of hair, beard, \etc, while acquiring detailed 3D geometries of these regions is challenging. Finally, they can directly synthesize accessories present in the source image, including eyeglasses, hats, and scarves, without their 3D models.

However, existing 2D-based one-shot talking-head methods~\cite{siarohin2019first,wang2019few,zakharov2020fast} come with their own set of limitations. Due to the absence of 3D graphics models, they can only synthesize the talking-head from the original viewpoint. They cannot render the talking-head from a novel view.

\begin{figure}[t!]
    \centering
    \vspace{-.1in}
    \includegraphics[width=.44\textwidth]{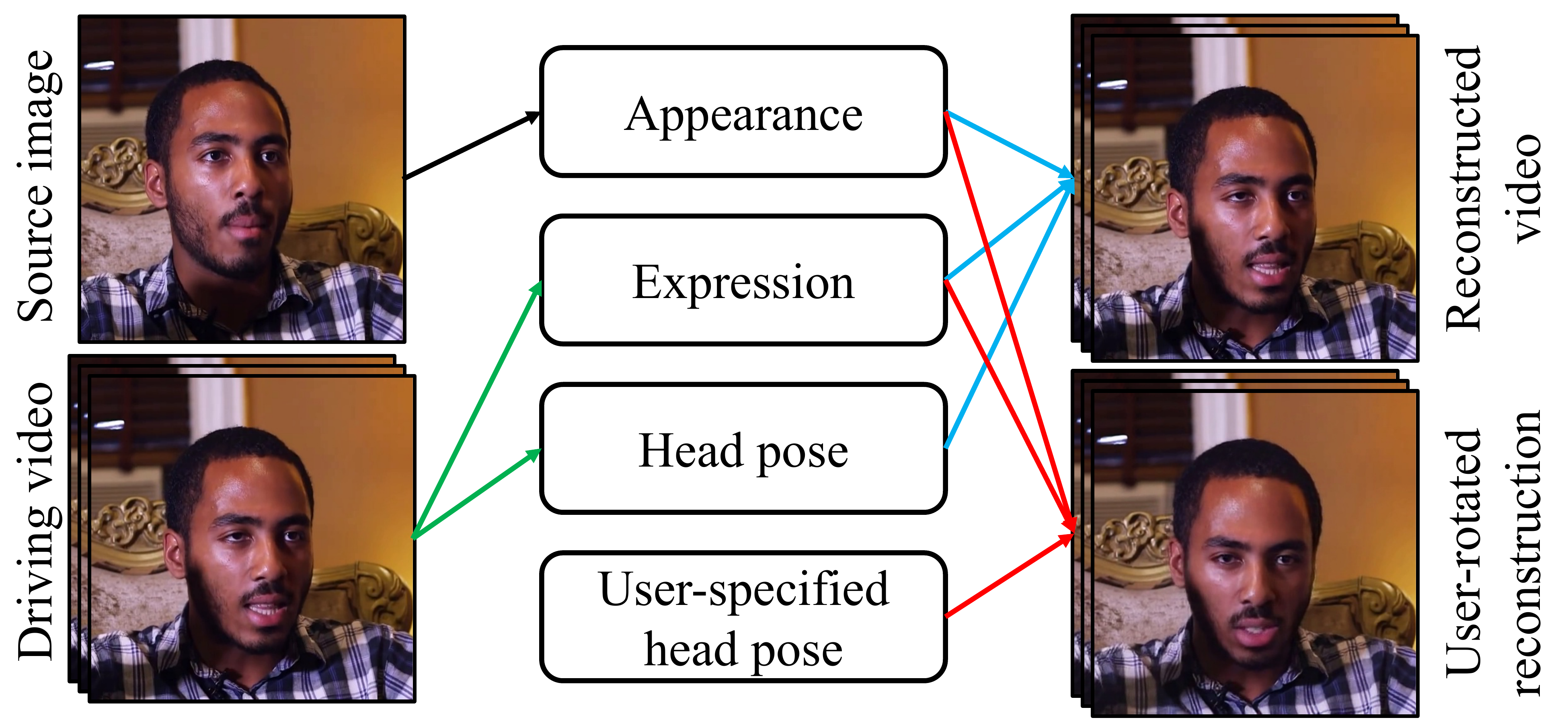} \vspace{-3mm}
    \caption{Combining appearance information from the source image, our framework can re-create a driving video by just using the expression and head pose information from the driving video. With a user-specified head pose, it can also synthesize the head pose change in the output video.}
    \label{fig:method_overview}
    \vspace{-.2in}
\end{figure}

Our approach addresses the fixed viewpoint limitation and achieves \textit{local free-view} synthesis. One can freely change the viewpoint of the talking-head within a large neighborhood of the original viewpoint, as shown in Fig.~\ref{figure:teaser}(c). Our model achieves this capability by representing a video using a novel 3D keypoint representation, where person-specific and motion-related information is decomposed. Both the keypoints and their decomposition are learned unsupervisedly. Using the decomposition, we can apply 3D transformations to the person-specific representation to simulate head pose changes such as rotating the talking-head in the output video.  Figure~\ref{fig:method_overview} gives an overview of our approach.

We conduct extensive experimental validation with comparisons to state-of-the-art methods. We evaluate our method on several talking-head synthesis tasks, including video reconstruction, motion transfer, and face redirection. We also show how our approach can be used to reduce the bandwidth of video conferencing, which has become an important platform for social networking and remote collaborations. By sending only the keypoint representation and reconstructing the source video on the receiver side, we can achieve a 10x bandwidth reduction as compared to the commercial H.264 standard without compromising the visual quality.

\noindent{\bf Contribution 1.} A novel one-shot neural talking-head synthesis approach, which achieves better visual quality than state-of-the-art methods on the benchmark datasets.

\noindent{\bf Contribution 2.} Local free-view control of the output video, without the need for a 3D graphics model. Our model allows changing the viewpoint of the talking-head during synthesis.

\noindent{\bf Contribution 3.} Reduction in bandwidth for video streaming. We compare our approach to the commercial H.264 standard on a benchmark talking-head dataset and show that our approach can achieve 10$\times$ bandwidth reduction.

\section{Related Works}
\noindent{\bf GANs.}
Since its introduction by Goodfellow~\etal~\cite{goodfellow2014generative}, GANs have shown promising results in various areas~\cite{liu2021generative}, such as unconditional image synthesis~\cite{goodfellow2014generative,radford2015unsupervised,liu2016coupled,gulrajani2017improved,karras2017progressive,karras2018style,karras2020analyzing}, image translation~\cite{isola2017image,taigman2016unsupervised,bousmalis2016unsupervised,shrivastava2016learning,zhu2017unpaired,liu2016unsupervised,huang2018multimodal,zhu2017toward,wang2018high,choi2017stargan,park2019semantic,liu2019few}, text-to-image translation~\cite{reed2016generative,zhang2017stackgan,xu2018attngan}, image processing~\cite{dong2015image,dong2016accelerating,kim2016accurate,ledig2017photo,lim2017enhanced,tong2017image,tai2017memnet,kupyn2018deblurgan,liu2018image, xiong2019foreground, zeng2019learning, iizuka2017globally}, and video synthesis~\cite{chan2019everybody,wang2018video,zhou2019dance,wang2019few,kim2018deep,pumarola2018ganimation,aberman2019deep,liu2019neural,siarohin2019monkeynet,liu2019liquid,ren2020deep}. We focus on using GANs to synthesize talking-head videos in this work.

\noindent{\bf 3D model-based talking-head synthesis.}
Works on transferring the facial motion of one person to another---face reenactment---can be divided into \emph{subject-dependent} and \emph{subject-agnostic} models. Traditional 3D-based methods usually build a subject-dependent model, which can only synthesize one subject. Moreover, they focus on transferring the expressions without the head movement~\cite{vlasic2005face,thies2015real,thies2016face2face,suwajanakorn2017synthesizing,thies2019deferred}. This line of works starts by collecting footage of the target person to be synthesized using an RGB or RGBD sensor~\cite{thies2016face2face,thies2015real}. Then a 3D model of the target person is built for the face region~\cite{blanz1999morphable}. At test time, the new expressions are used to drive the 3D model to generate the desired motions.

More recent 3D model-based methods are able to perform subject-agnostic face synthesis~\cite{olszewski2017realistic, nagano2018pagan, geng2018warp, fried2019text}. While they can do an excellent job synthesizing the inner face region, they have a hard time generating realistic hair, teeth, accessories, etc. Due to the limitations, most modern face reenactment frameworks adopt the 2D approach. Another line of works~\cite{deng2020disentangled, tewari2020stylerig} focuses on controllable face generation, providing explicit control over the generated face from a pretrained StyleGAN~\cite{karras2018style,karras2020analyzing}. However, it is not clear how they can be adapted to modifying real images since the inverse mapping from images to latent codes is nontrivial.

\noindent{\bf 2D-based talking-head synthesis.}
Again, 2D approaches can be classified into subject-dependent and subject-agnostic models. Subject-dependent models~\cite{bansal2018recycle, wu2018reenactgan} can only work on specific persons since the model is only trained on the target person. On the other hand, subject-agnostic models~\cite{averbuch2017bringing, wiles2018x2face, pumarola2018ganimation, geng2018warp, zhou2019talking, chen2019hierarchical, song2019talking, jamaludin2019you, vougioukas2019realistic, nirkin2019fsgan, zakharov2019few, wang2019few, siarohin2019first, gu2020flnet, burkov2020neural, zakharov2020fast, ha2020marionette} only need a single image of the target person, who is not seen during training, to synthesize arbitrary motions. Siarohin~\etal~\cite{siarohin2019first} warp extracted features from the input image, using motion fields estimated from sparse keypoints. On the other hand, Zakharov~\etal~\cite{zakharov2019few} demonstrate that it is possible to achieve promising results using direct synthesis methods without any warping. Few-shot vid2vid~\cite{wang2019few} injects the information into their generator by dynamically determining the the parameters in the SPADE~\cite{park2019semantic} modules. Zakharov~\etal~\cite{zakharov2020fast} decompose the low and high frequency components of the image and greatly accelerate the inference speed of the network. While demonstrating excellent result qualities, these methods can only synthesize fixed viewpoint videos, which produce less immersive experiences.

\noindent{\bf Video compression.}
A number of recent works~\cite{agustsson2020scale,djelouah2019neural,hu2020improving,lin2020m,lu2019dvc,rippel2019learned,wu2018video} propose using a deep network to compress arbitrary videos. The general idea is to treat the problem of video compression as one of interpolating between two neighboring keyframes. Through the use of deep networks to replace various parts of the traditional pipeline, as well as techniques such as hierarchical interpolation and joint encoding of residuals and optical flows, these prior works reduce the required bit-rate. Other works~\cite{lu2018deep,wang2019edvr,yang2018multi,zhang2020davd} focus on restoring the quality of low bit-rate videos using deep networks. Most related to our work is DAVD-Net~\cite{zhang2020davd}, which restores talking-head videos using information from the audio stream.
Our proposed method is different from these works in a number of aspects, in both the goal as well as the method used to achieve compression. We specifically focus on videos of talking faces. People's faces have an inherent structure---from the shape to the relative arrangement of different parts such as eyes, nose, mouth, \etc. This allows us to use keypoints and associated metadata for efficient compression, an order of magnitude better than traditional codecs. Our method does not guarantee pixel-aligned output videos; however, it faithfully models facial movements and emotions. It is also better suited for video streaming as it does not use bi-directional or B-frames.

\section{Method}
\label{sec:method}

\begin{figure}[t!]
    \centering
    \vspace{-.1in}
    \includegraphics[width=.45\textwidth]{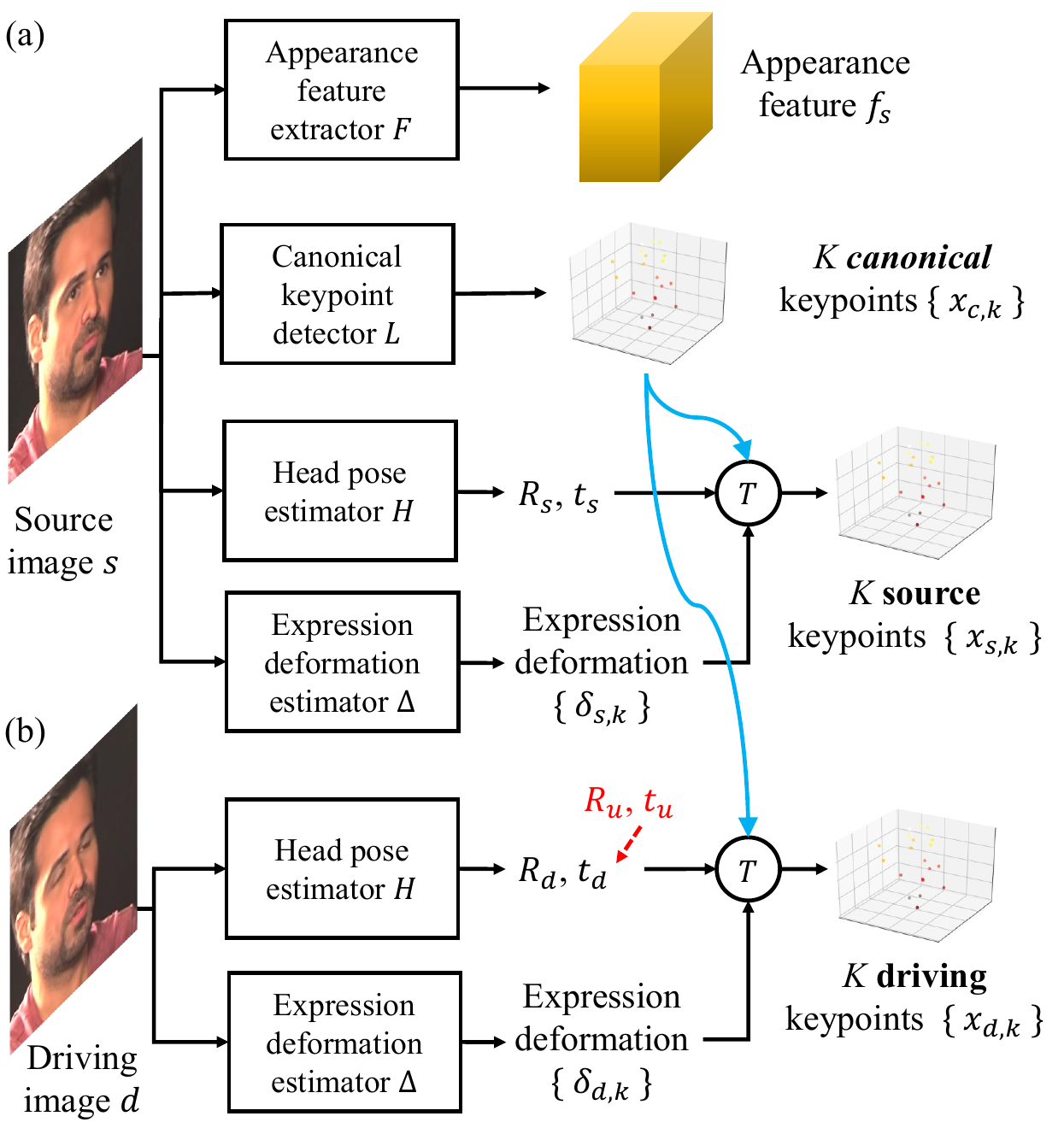}
    \vspace{-2mm}
    \caption{Source and driving feature extraction. (a) From the source image, we extract appearance features and 3D canonical keypoints. We also estimate the head pose and the keypoint perturbations due to expressions. We use them to compute the source keypoints. (b) For the driving image, we again estimate the head pose and the expression deformations. By reusing canonical keypoints from the source image, we compute the driving keypoints. 
    }
    \label{fig:method12_overview}
    \vspace{-.1in}
\end{figure}
\begin{figure*}[tbh!]
    \centering
    \vspace{-.1in}
    \includegraphics[width=.9\textwidth]{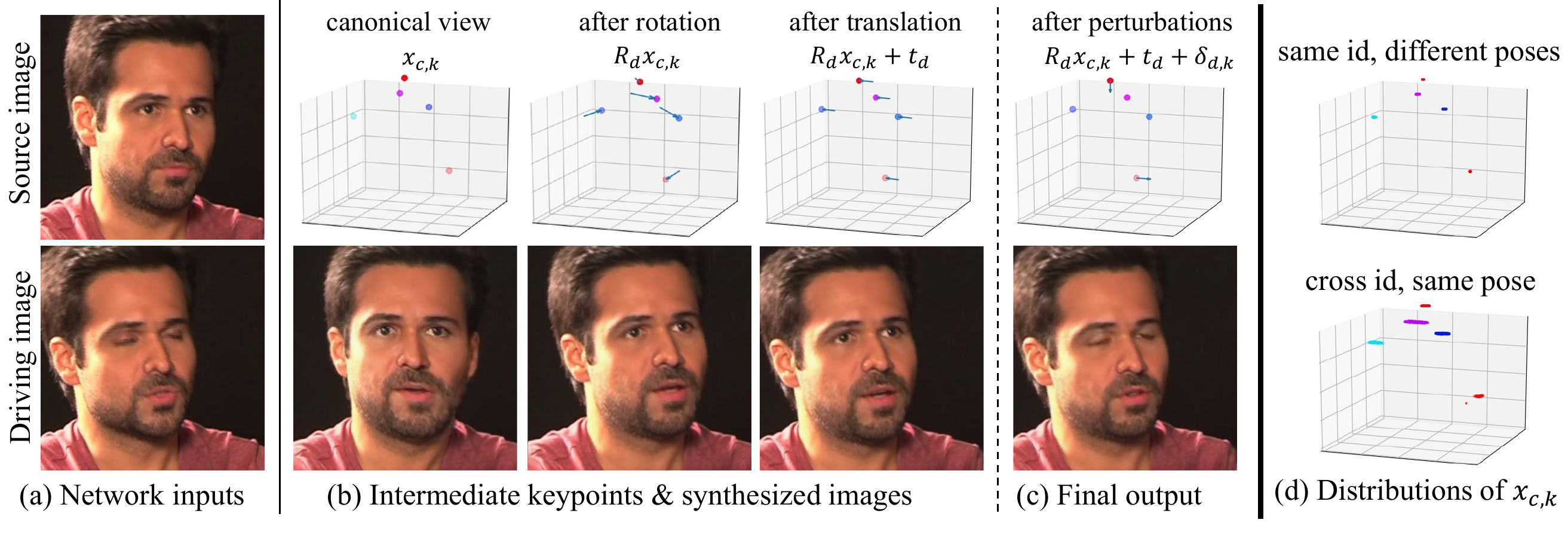}
    \vspace{-.1in}
    \caption{Keypoint computation pipeline. For each step, we show the first five keypoints and the synthesized images using them. Given the source image (a), our model first predicts the canonical keypoints (b). We then apply the rotation and translation estimated from the driving image to the canonical keypoints, bringing them to the target head pose (transformations illustrated as arrows). (c) The expression-aware deformation adjusts the keypoints to the target expression (\eg closed eyes). (d) We visualize the distributions of canonical keypoints estimated from different images. Upper: the canonical keypoints from different poses of a person are similar. Lower: the canonical keypoints from different people in the same pose are different.}
    \label{fig:intermediate}

	\vspace{3mm}
	\centering
	\includegraphics[width=.95\textwidth]{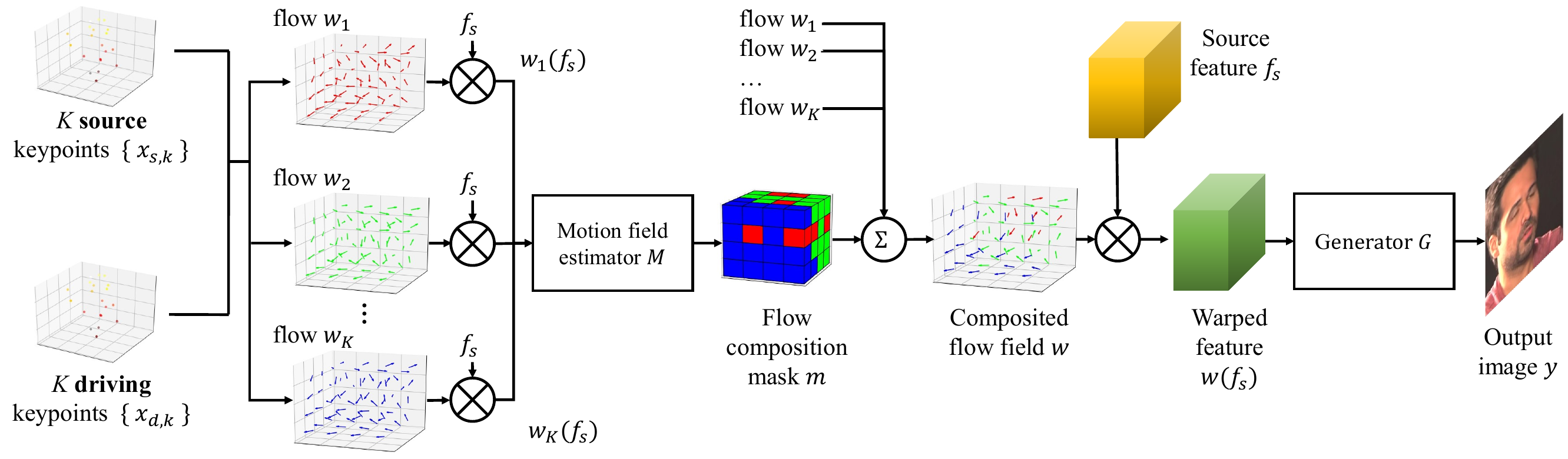}
	\vspace{-.05in}
	\caption{Video synthesis. We use the source and driving keypoints to estimate $K$ flows, $w_k$'s. These flows are used to warp the source feature $f_s$. The results are combined and fed to the motion field estimation network $M$ to produce a flow composition mask $m$. A linear combination of $m$ and $w_k$'s then produces the composited flow field $w$, which is used to warp the 3D source feature. Finally, the generator $G$ converts the warped feature to the output image $y$.}
	\label{fig:method3_overview}
    \vspace{-.1in}
\end{figure*}

Let $s$ be an image of a person, referred to as the source image. Let $\{d_1, d_2, ..., d_{N}\}$ be a talking-head video, called the driving video, where $d_i$'s are the individual frames, and $N$ is the total number of frames. Our goal is to generate an output video $\{y_1, y_2, ...,y_{N}\}$, where the identity in $y_i$'s is inherited from $s$ and the motions are derived from $d_i$'s. Several talking-head synthesis tasks fall in the above setup. When $s$ is a frame of the driving video (\eg, the first frame: $s\equiv d_1$.), we have a video reconstruction task. When $s$ is not from the driving video, we have a motion transfer task.

We propose a pure neural synthesis approach that does not use any 3D graphics models, such as the well-known 3D morphable model (3DMM)~\cite{blanz1999morphable}. Our approach contains three major steps: \textit{1) source image feature extraction}, \textit{2) driving video feature extraction}, and \textit{3) video generation}. In Fig.~\ref{fig:method12_overview}, we illustrate \textit{1)} and \textit{2)}, while Fig.~\ref{fig:method3_overview} shows \textit{3)}. Our key ingredient is an unsupervised approach for learning a set of \textbf{3D} keypoints and their decomposition. We decompose the keypoints into two parts, one that models the facial expressions and the other that models the geometric signature of a person. These two parts are combined with the target head pose to generate the image-specific keypoints. After the keypoints are estimated, they are then used to learn a mapping function between two images. We implement these steps using a set of networks and train them jointly. In the following, we discuss the three steps in detail.

\subsection{Source image feature extraction}

Synthesizing a talking-head requires knowing the appearance of the person, such as the skin and eye colors. As shown in Fig.~\ref{fig:method12_overview}(a), we first apply a 3D appearance feature extraction network $F$ to map the source image $s$ to a 3D appearance feature volume $f_{s}$. Unlike a 2D feature map, $f_{s}$ has three spatial dimensions: width, height, and depth. Mapping to a 3D feature volume is a crucial step in our approach. It allows us to operate the keypoints in the 3D space for rotating and translating the talking-head during synthesis.

We extract a set of $K$ 3D keypoints $x_{c,k} \in \mathbb{R}^3$ from $s$ using a canonical 3D keypoint detection network $L$. We set $K=20$ throughout the paper unless specified otherwise. Note that these keypoints are unsupervisedly learned and different from the common facial landmarks. We note that the extracted keypoints are meant to be independent of the face's pose and expression. They shall only encode a person's geometry signature in a neutral pose and expression.

Next, we extract pose and expression information from the image. We use a head pose estimation network $H$ to estimate the head pose of the person in $s$, parameterized by a rotation matrix $R_s \in \mathbb{R}^{3\times 3}$ and a translation vector $t_s \in \mathbb{R}^{3}$. In addition, we use an expression deformation estimation network~$\Delta$ to estimate a set of $K$ 3D deformations $\delta_{s,k}$---the deformations of keypoints from the neutral expression. Both $H$ and $\Delta$ extract motion-related geometry information in the image. We combine the identity-specific information extracted by $L$ with the motion-related information extracted by $H$ and $\Delta$ to obtain the source 3D keypoints $x_{s,k}$ via a transformation $T$:
\begin{align}
x_{s,k}
&=T(x_{c,k}, R_s, t_s, \delta_{s,k})\equiv R_s x_{c,k} + t_s + \delta_{s,k}\label{eqn:decomposition}
\end{align}
The final keypoints are image-specific and contain person-signature, pose, and expression information. Figure~\ref{fig:intermediate} visualizes the keypoint computation pipeline.

The 3D keypoint decomposition in~(\ref{eqn:decomposition}) is of paramount importance to our approach. It commits to a prior decomposition of keypoints: geometry-signatures, head poses, and expressions. It helps learn manipulable representations and differs our approach from prior 2D keypoint-based neural talking-head synthesis approaches~\cite{siarohin2019first,wang2019few,zakharov2020fast}.
Also note that unlike FOMM~\cite{siarohin2019first}, our model does not estimate Jacobians. The Jacobian represents how a local patch around the keypoint can be transformed into the corresponding patch in another image via an affine transformation. Instead of explicitly estimating them, our model assumes the head is mostly rigid and the local patch transformation can be directly derived from the head rotation via $J_s = R_s$. Avoiding estimating Jacobians allows us to further reduce the transmission bandwidth for the video conferencing application, as detailed in Sec.~\ref{sec:compression}.

\subsection{Driving video feature extraction}

We use $d$ to denote a frame in $\{d_1, d_2, ..., d_N\}$ as individual frames are processed in the same way. To extract motion-related information, we apply the head pose estimator $H$ to get $R_d$ and $t_d$ and apply the expression deformation estimator $\Delta$ to obtain $\delta_{d,k}$'s, as shown in Fig.~\ref{fig:method12_overview}(b).

Now, instead of extracting canonical 3D keypoints from the driving image $d$ using $L$, we reuse $x_{c,k}$, which were extracted from the source image $s$. This is because the face in the \textit{output} image must have the same identity as the one in the source image $s$. There is no need to compute them again. Finally, the identity-specific information and the motion-related information are combined to compute the driving keypoints for the driving image $d$ in the same way we obtained source keypoints:
\begin{align}
x_{d,k}
&=T(x_{c,k}, R_d, t_d, \delta_{d,k})=R_d x_{c,k} + t_d + \delta_{d,k}
\end{align}
We apply this processing to each frame in the driving video, and each frame can be compactly represented by $R_d$, $t_d$, and $\delta_{d,k}$'s. This compact representation is very useful for low-bandwidth video conferencing. In Sec.~\ref{sec:compression}, we will introduce an entropy coding scheme to further compress these quantities to reduce the bandwidth utilization.

Our approach allows manual changes to the 3D head pose during synthesis. Let $R_{u}$ and $t_{u}$ be user-specified rotation and translation, respectively. The final head pose in the output image is given by $R_d \leftarrow R_u R_{d}$ and $t_d \leftarrow t_u + t_{d}$. In video conferencing, we can change a person's head pose in the video stream freely despite the original view angle.

\subsection{Video generation}

As shown in Fig.~\ref{fig:method3_overview}, we synthesize an output image by warping the source feature volume and then feeding the result to the image generator $G$ to produce the output image $y$. The warping approximates the nonlinear transformation from $s$ to $d$. It \textit{re-positions} the source features for the synthesis task.

To obtain the required warping function $w$, we take a bottom-up approach. We first compute the warping flow $w_k$ induced by the $k$-th keypoint using the first order approximation~\cite{siarohin2019first}, which is reliable only around the neighborhood of the keypoint. After obtaining all $K$ warping flows, we apply each of them to warp the source feature volume. The $K$ warped features are aggregated to estimate a flow composition mask $m$ using the motion field estimation network $M$. This mask indicates which of the $K$ flows to use at each spatial 3D location. We use this mask to combine the $K$ flows to produce the final flow $w$. Details of the operation are given in \supplementary{A.1}.
\ifdefined\cameraReady 
(For all appendices, please refer to our \href{https://arxiv.org/abs/2011.15126}{full technical report}~\cite{wang2020facevid2vid}.)
\fi

\subsection{Training} 

We train our model using a dataset of talking-head videos where each video contains a single person. For each video, we sample two frames: one as the source image $s$ and the other as the driving image $d$. We train the networks $F$, $\Delta$, $H$, $L$, $M$, and $G$ by minimizing the following loss:
\setlength{\belowdisplayskip}{5pt} \setlength{\belowdisplayshortskip}{5pt}
\setlength{\abovedisplayskip}{5pt} \setlength{\abovedisplayshortskip}{5pt}
\begin{align}
&\mathcal{L}_{P}(d, y) + \mathcal{L}_{G}(d, y) +
\mathcal{L}_{E}(
\{x_{d,k}\}) +
\nonumber\\
&\mathcal{L}_{L}(\{ x_{d,k}\}) +
\mathcal{L}_{H}(R_d, \bar{R}_d) +
\mathcal{L}_{\Delta}(\{\delta_{d,k}\})
\end{align}
In short, the first two terms ensure the output image looks similar to the ground truth. The next two terms enforce the predicted keypoints to be consistent and satisfy some prior knowledge about the keypoints. The last two terms constrain the estimated head pose and keypoint perturbations. We briefly discuss these losses below and leave the implementation details in \supplementary{A.2}.

\noindent{\bf Perceptual loss $\mathcal{L}_P$.} We minimize the perceptual loss~\cite{johnson2016perceptual,wang2018high} between the output and the driving image, which is helpful in producing sharp-looking outputs.

\noindent{\bf GAN loss $\mathcal{L}_{G}$.} We use a multi-resolution patch GAN where the discriminator predicts at the patch-level. We also minimize the discriminator feature matching loss~\cite{wang2018high,wang2019few}.

\noindent{\bf Equivariance loss $\mathcal{L}_E$.} This loss ensures the consistency of image-specific keypoints ${x}_{d,k}$. For a valid keypoint, when applying a 2D transformation to the image, the predicted keypoints should change according to the applied transformation~\cite{zhang2018unsupervised,siarohin2019first}. Since we predict 3D instead of 2D keypoints, We use an orthographic projection to project the keypoints to the image plane before computing the loss.

\noindent{\bf Keypoint prior loss $\mathcal{L}_L$.} We use a keypoint coverage loss to encourage the estimated image-specific keypoints ${x}_{d,k}$'s to spread out across the face region, instead of crowding around a small neighborhood. We compute the distance between each pair of the keypoints and penalize the model if the distance falls below a preset threshold. We also use a keypoint depth prior loss that encourages the mean depth of the keypoints to be around a preset value.

\noindent{\bf Head pose loss $\mathcal{L}_H$.} We penalize the prediction error of the head rotation $R_d$ compared to the ground truth $\bar{R}_d$. Since acquiring the ground truth head pose for a large-scale video dataset is expensive, we use a pre-trained pose estimation network~\cite{ruiz2018fine} to approximate $\bar{R}_d$.

\noindent{\bf Deformation prior loss $\mathcal{L}_{\Delta}$.} The loss penalizes the magnitude of the deformations $\delta_{d,k}$'s. As the deformations model the deviation from the canonical keypoints due to expression changes, their magnitudes should be small. 

\vspace{-0.05in}
\section{Experiments}
\vspace{-0.05in}

\noindent{\bf Implementation details.} The network architecture and training hyper-parameters are available in \supplementary{A.3}. 

\noindent{\bf Datasets.} Our evaluation is based on VoxCeleb2~\cite{chung2018voxceleb2} and TalkingHead-1KH, a newly collected large-scale talking-head video dataset. It contains $180$K videos, which are often with higher quality and larger resolution than those in VoxCeleb2. Details are available in \supplementary{B.1}.

\subsection{Talking-head image synthesis}
\begin{table*}[tb!]
    \vspace{-.15in}
	\caption{Comparisons with state-of-the-art methods on face reconstruction. $\uparrow$ larger is better. $\downarrow$ smaller is better.}\label{tbl::comparison}\vspace{-3mm}
	\centering
	\small
	\setlength{\tabcolsep}{4pt}
    \begin{tabular}{rcccccccccccc}
\toprule
& \multicolumn{6}{c}{VoxCeleb2~\cite{chung2018voxceleb2}} & \multicolumn{6}{c}{TalkingHead-1KH}  \\
\small
Method & L1$\downarrow$ & PSNR$\uparrow$ & SSIM$\uparrow$ & MS-SSIM$\uparrow$ & FID$\downarrow$ & AKD$\downarrow$ & L1$\downarrow$ & PSNR$\uparrow$ & SSIM$\uparrow$ & MS-SSIM$\uparrow$ & FID$\downarrow$ & AKD$\downarrow$ \\
\cmidrule(r){2-7}
\cmidrule(r){8-13}
fs-vid2vid~\cite{wang2019few} &
17.10 & 20.36 & 0.71 & Nan & 85.76 & 3.41 & 
15.18 & 20.94 & 0.75 & Nan & 63.47 & 11.07 \\
FOMM~\cite{siarohin2019first}  &
12.66 & 23.25 & 0.77 & 0.83 & 73.71 & 2.14 &
12.30 & 23.67 & 0.79 & 0.83 & 55.35 & 3.76 \\
FOMM-L~\cite{siarohin2019first} &
N/A & N/A & N/A & N/A & N/A & N/A  &
12.81 & 23.13 & 0.78 & Nan & 60.58 & 4.04 \\
Bi-layer~\cite{zakharov2020fast} & 23.95 & 16.98 & 0.66 & 0.66 & 203.36 & 5.38 &
N/A & N/A & N/A & N/A & N/A & N/A \\
{\bf Ours } & 
{\bf 10.74} & {\bf 24.37} & {\bf 0.80} & {\bf 0.85} & {\bf 69.13} & {\bf 2.07} &
{\bf 10.67} & {\bf 24.20} & {\bf 0.81} & {\bf 0.84} & {\bf 52.08} & {\bf 3.74} \\
\bottomrule
	\end{tabular}
	\vspace{-.1in}
 \end{table*}
\begin{figure*}[t!]
    \centering
    \includegraphics[width=.9\textwidth]{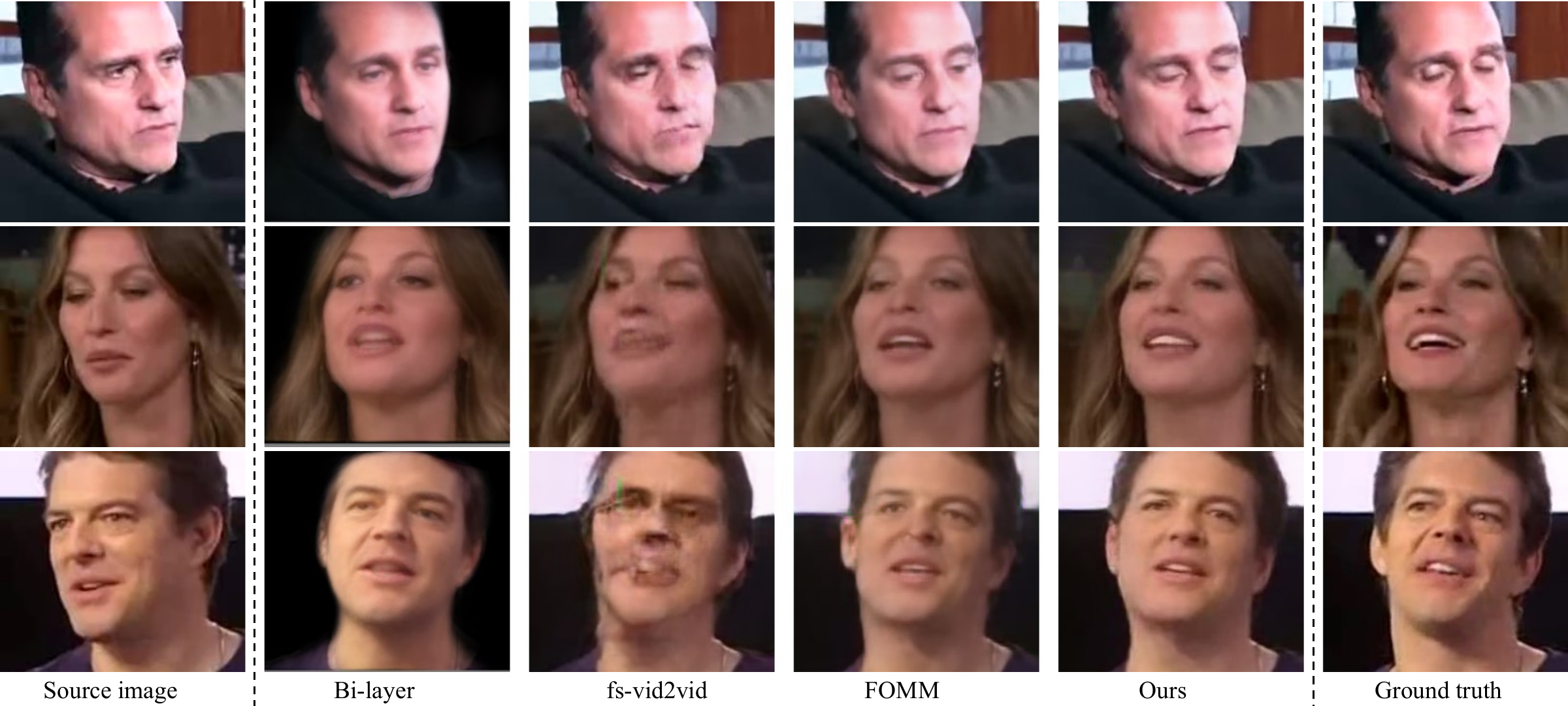}
    \vspace{-.1in}
    \caption{Qualitative comparisons on the Voxceleb2 dataset~\cite{chung2018voxceleb2}. Our method better captures the driving motions.}
    \label{fig:result_voxceleb}
    
    \centering
    \vspace{2mm}
    \includegraphics[width=.9\textwidth]{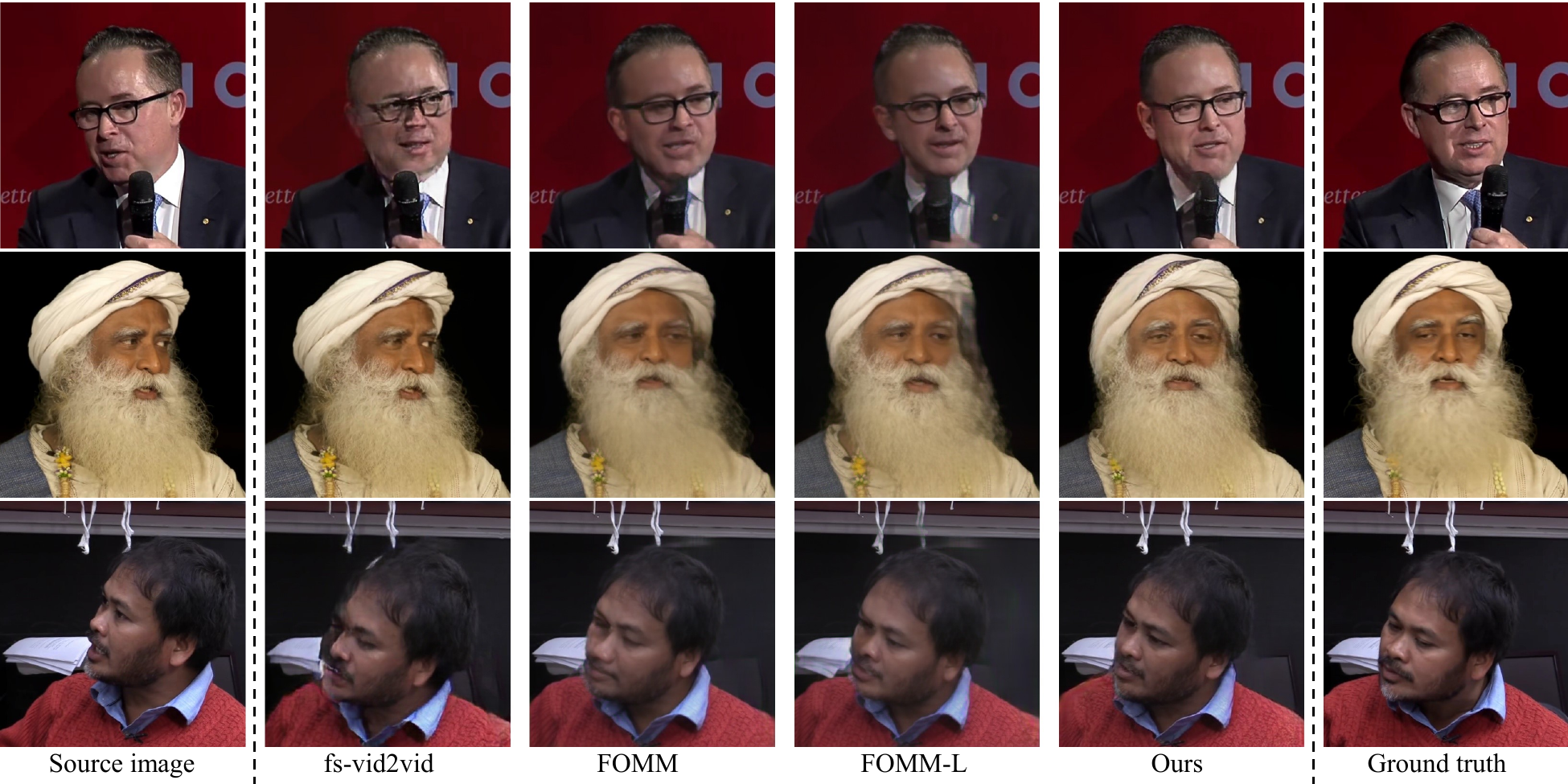}
    \vspace{-.1in}
    \caption{Qualitative comparisons on the TalkingHead-1KH dataset. Our method produces more faithful and sharper results.}
    \label{fig:result_youtube}
    \vspace{-6mm}
\end{figure*}

\noindent{\bf Baselines.} We compare our neural talking-head model with three state-of-the-art methods: FOMM~\cite{siarohin2019first}, few-shot vid2vid (fs-vid2vid)~\cite{wang2019few}, and bi-layer neural avatars (bi-layer)~\cite{zakharov2020fast}. We use the released pre-trained model on VoxCeleb2 for bi-layer~\cite{zakharov2020fast}, and retrain from scratch for others on the corresponding datasets. Since bi-layer does not predict the background, we subtract the background when doing quantitative analyses.

\noindent{\bf Metrics.} We evaluate a synthesis model on 1) reconstruction faithfulness using $L_1$, PSNR, SSIM/MS-SSIM, 2) output visual quality using FID, and 3) semantic consistency using average keypoint distance (AKD). Please consult \supplementary{B.2}~for details of the performance metrics.

\noindent{\bf Same-identity reconstruction.}
We first compare the face synthesis results where the source and driving images are of the same person. The quantitative evaluation is shown in Table~\ref{tbl::comparison}. It can be seen that our method outperforms other competing methods on all metrics for both datasets. To verify that our superior performance does not come from more parameters, we train another large FOMM model with doubled filter size (FOMM-L), which is larger than our model. We can see that enlarging the model actually hurts the performance, proving that simply making the model larger does not help. Figures~\ref{fig:result_voxceleb} and~\ref{fig:result_youtube} show the qualitative comparisons. Our method can more faithfully reproduce the driving motions.

\begin{figure}[t!]
    \vspace{-.25in}
    \centering
    \includegraphics[width=\columnwidth]{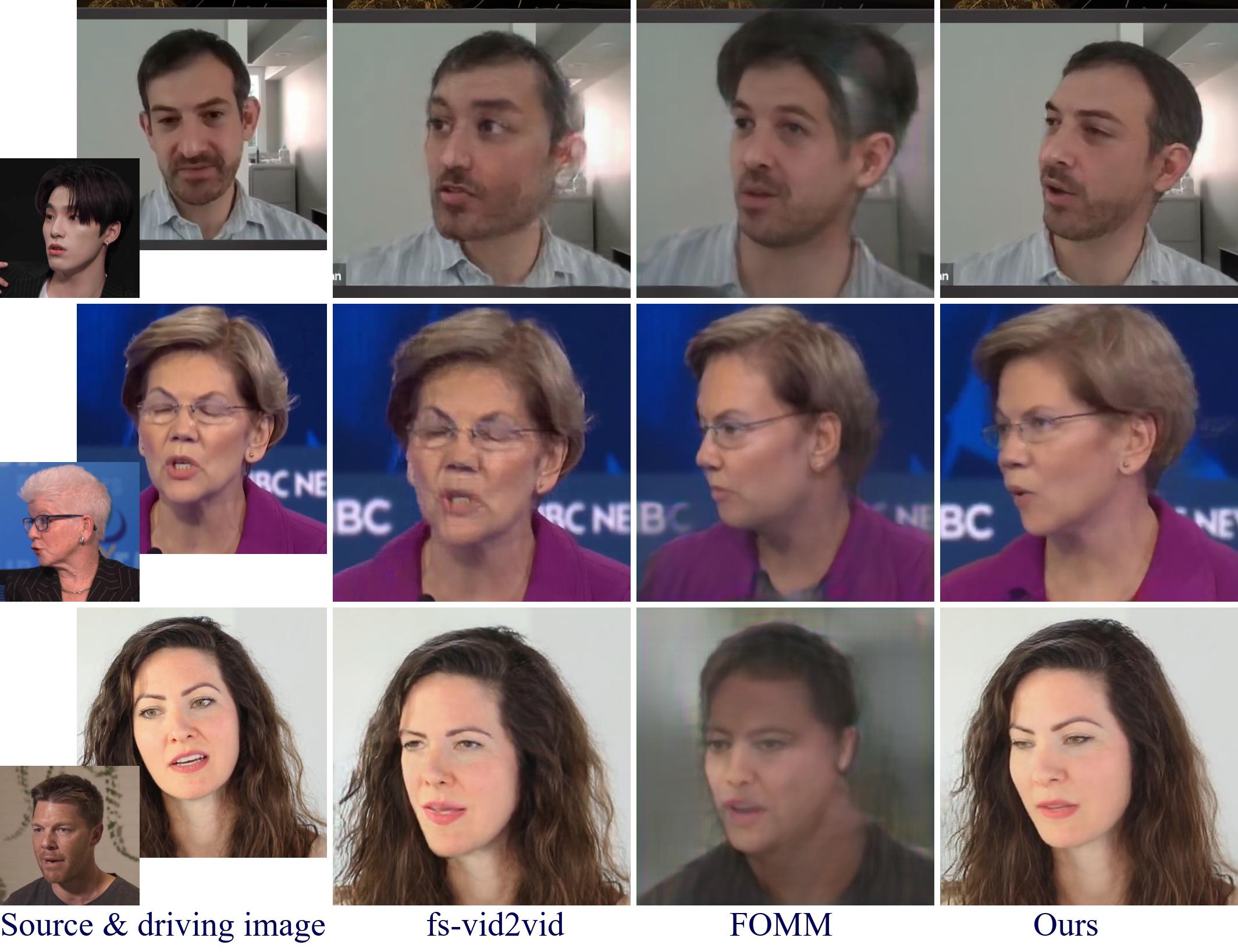}
    \vspace{-.3in}
    \caption{Qualitative results for cross-subject motion transfer. Ours captures the motion and preserves the identity better.}
    \label{fig:result_motion_transfer}
    \vspace{2mm}

    \centering
    \includegraphics[width=\columnwidth]{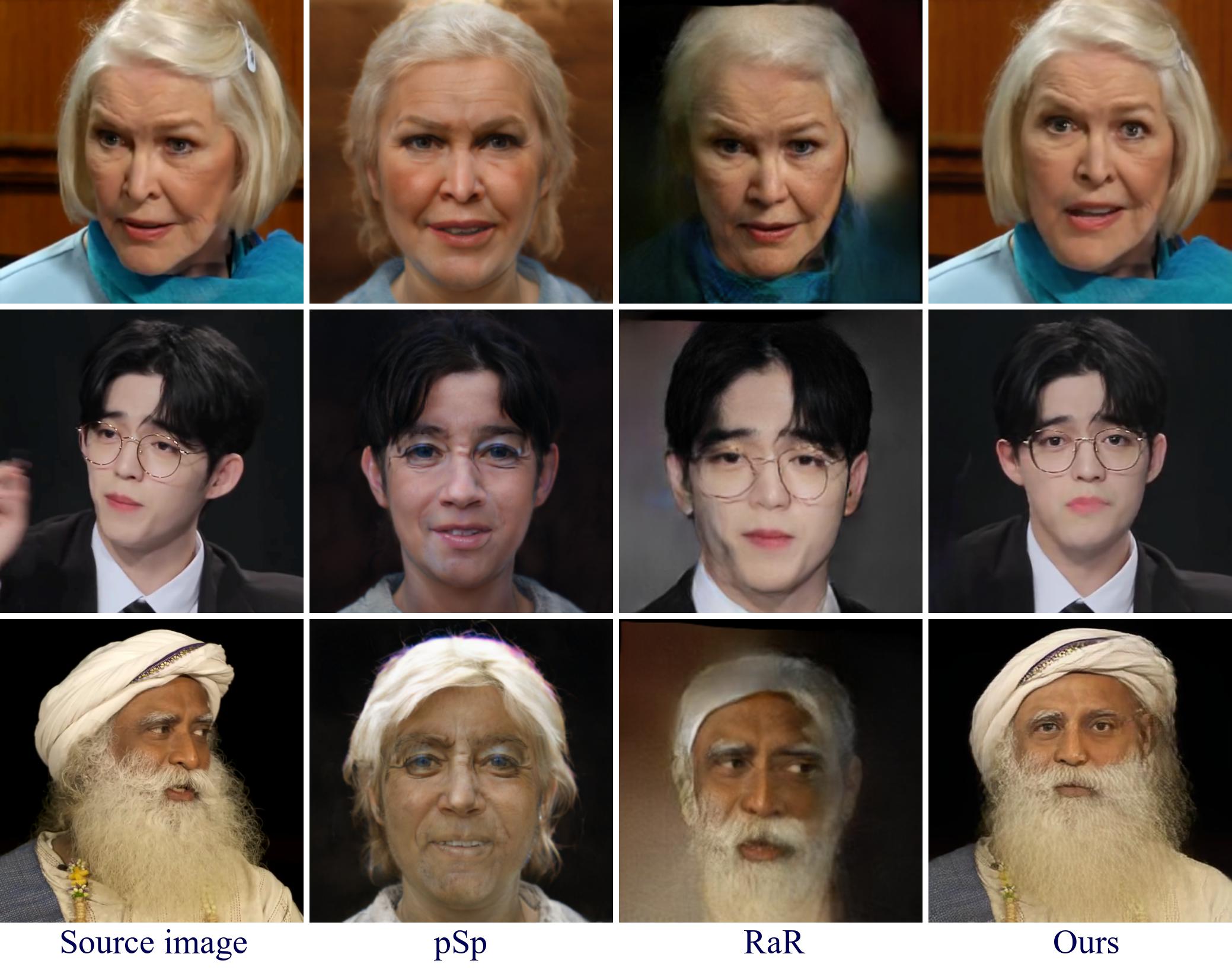}
    \vspace{-.3in}
    \caption{Qualitative results for face frontalization. Our method more realistically frontalizes the faces.}
    \label{fig:result_frontalize}
    \vspace{-.2in}
\end{figure}

\noindent{\bf Cross-identity motion transfer.}
Next, we compare results where the source and driving images are from different persons (cross-identity). Table~\ref{tbl::relative} shows that our method achieves the best results compared to other methods. Figure~\ref{fig:result_motion_transfer} compares results from different approaches. It can be seen that our method generates more realistic images while still preserving the original identity. For cross-identity motion transfer, it is sometimes useful to use relative motion~\cite{siarohin2019first}, where only motion differences between two neighboring frames in the driving video are transferred. We report comparisons using relative motion in \supplementary{B.3}.

\noindent{\bf Ablation study.} We benchmark the performance gains from the proposed keypoint decomposition scheme, the mask estimation network, and pose supervision in \supplementary{B.4}.

\noindent{\bf Failure cases.} Our model fails when large occlusions and image degradation occur, as visualized in \supplementary{B.5}.

\noindent{\bf Face recognition.} Since the canonical keypoints are independent of poses and expressions, they can also be applied to face recognition. In \supplementary{B.6}, we show that this achieves 5x accuracy than using facial landmarks.

\begin{table}[tb!]
    \vspace{-.1in}
	\caption{Quantitative results on cross-identity motion transfer. Our method achieves lowest FIDs and highest identity-preserving scores (CSIM~\cite{zakharov2019few}).}
\label{tbl::relative}\vspace{-3mm}
	\centering
	\small
	\setlength{\tabcolsep}{4pt}
    \begin{tabular}{rcccc}
\toprule
& \multicolumn{2}{c}{VoxCeleb2~\cite{chung2018voxceleb2}} & \multicolumn{2}{c}{TalkingHead-1KH} \\
Method & FID$\downarrow$ & CSIM$\uparrow$ & FID$\downarrow$ & CSIM$\uparrow$ \\
\cmidrule(r){2-3}
\cmidrule(r){4-5}
fs-vid2vid~\cite{wang2019few} & 59.84 & 0.593 & 52.72 & 0.703 \\
FOMM~\cite{siarohin2019first} & 84.06 & 0.582 & 87.32 & 0.542\\
{\bf Ours } & {\bf 55.64} & {\bf 0.753} & {\bf 46.99} & {\bf 0.777}\\
\bottomrule
	\end{tabular}
	\vspace{1mm}
	
	\caption{Face frontalization quantitative comparisons. We compute the identity loss and angle difference for each method and report the percentage where the losses are within a threshold (0.05 and 15 degrees, respectively).}
	\label{tbl::rotation}

	\centering
	\small
	\setlength{\tabcolsep}{4pt}	
	\begin{tabular}{rcccc}
		\toprule
		Method & Identity (\%)$\uparrow$ & Angle (\%)$\uparrow$ & Both (\%)$\uparrow$ & FID$\downarrow$ \\
		\cline{1-5}
		pSp~\cite{richardson2020encoding} & 57.3 & {\bf 99.8} & 57.3 & 118.08 \\
		RaR~\cite{zhou2020rotate} & 55.1 & 87.2 & 50.8 & 78.81 \\
		{\bf Ours } & {\bf 94.3} & 90.9 & {\bf 85.9} & {\bf 23.87} \\
		\bottomrule
	\end{tabular}
	\vspace{-.15in}
\end{table}

\subsection{Face redirection.}

\noindent{\bf Baselines.} 
We benchmark our talking-head model's face redirection capability using latest face frontalization methods: pixel2style2pixel (pSp)~\cite{richardson2020encoding} and Rotate-and-Render (RaR)~\cite{zhou2020rotate}. pSp projects the original image into a latent code and then uses a pre-trained StyleGAN~\cite{abdal2019image2stylegan} to synthesize the frontalized image. RaR adopts a 3D face model to rotate the input image and re-renders it in a different pose.

\noindent{\bf Metrics.}
The results are evaluated by two metrics: identity preservation and head pose angles. We use a pre-trained face recognition network~\cite{parkhi2015deep} to extract high-level features, and compute the distance between the rotated face and the original one. We use a pre-trained head pose estimator~\cite{ruiz2018fine} to obtain head angles of the rotated face. For a rotated image, if its identity distance to the original image is within some threshold, and/or its head angle is within some tolerance to the desired angle, we consider it as a ``good" image. 

We report the ratio of ``good" images using our metric for each method in Table~\ref{tbl::rotation}. Example comparisons can be found in Fig.~\ref{fig:result_frontalize}. It can be seen that while pSp can always frontalize the face, the identity is usually lost. RaR generates more visually appealing results since it adopts 3D face models, but has problems outside the inner face regions. Besides, both methods have issues regarding the temporal stability. Only our method can realistically frontalize the inputs.

\vspace{-.1in}
\section{Neural Talking-Head Video Conferencing}\label{sec:compression}
\vspace{-.05in}

Our talking-head synthesis model distills motions in a driving image using a compact representation, as discussed in Sec.~\ref{sec:method}. Due to this advantage, our model can help reduce the bandwidth consumed by video conferencing applications. We can view the process of video conferencing as the receiver watching an animated version of the sender's face.

\begin{figure}[!t]
    \vspace{-.1in}
    \centering
    \includegraphics[width=.44\textwidth]{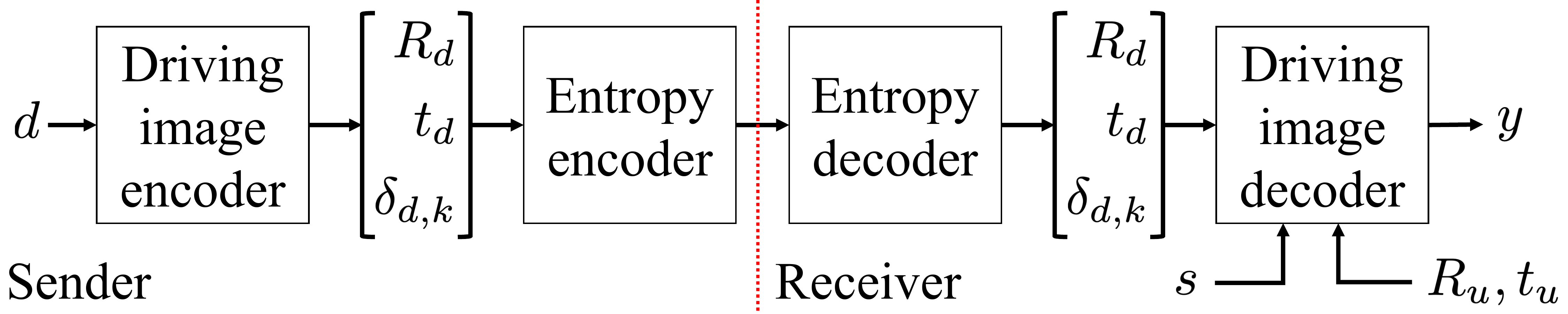}
    \vspace{-0.1in}
    \caption{Our video compression framework. On the sender's side, the driving image encoder extracts keypoint perturbations $\delta_{d,k}$ and head poses $R_d$ and $t_d$. They are then compressed using an entropy encoder and sent to the receiver. The receiver decompresses the message and uses them along with the source image $s$ to generate $y$, a reconstruction of the input $d$. Our framework can also change the head pose on the receiver's side by using the pose offset $R_u$ and $t_u$.}
    \label{fig:compression}\vspace{1mm}
    \centering
    \includegraphics[width=\columnwidth]{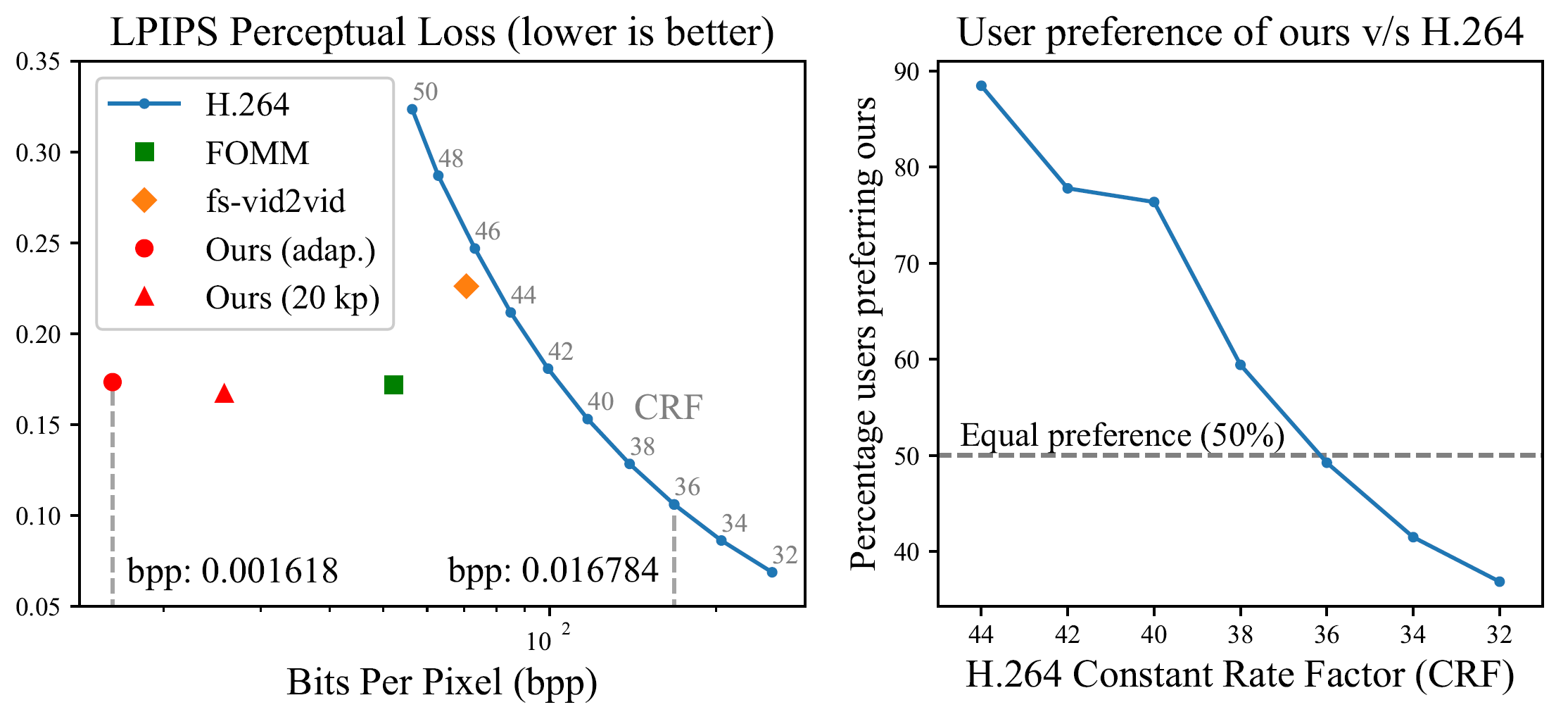}
    \vspace{-.3in}
    \caption{Automatic and human evaluations for video compression. Ours requires much lower bandwidth due to our keypoint decomposition and adaptive scheme.}
    \label{fig:compression_metrics}
    \vspace{-.2in}
\end{figure}

Figure~\ref{fig:compression} shows a video conferencing system built using our neural talking-head model. For a driving image $d$, we use the driving image encoder, consisting of $\Delta$ and $H$,
to extract the expression deformations~$\delta_{d,k}$ and the head pose $R_d, t_d$. By representing a rotation matrix using Euler angles, we have a compact representation of $d$ using $3K+6$ numbers: 3 for the rotation, 3 for the translation, and 3K for the deformations. We further compress these values using an entropy encoder~\cite{cover1999elements}. Details are in \supplementary{C.1}.

The receiver receives the entropy-encoded representation and uses the entropy decoder to recover $\delta_{d,k}$ and $R_d, t_d$. They are then fed into our talking-head synthesis framework to reconstruct the original image $d$. We assume that the source image $s$ is sent to the receiver at the beginning of the video conferencing session or re-used from a previous session. Hence, it does not consume additional bandwidth. We note that the source image is different from the I-frame in traditional video codecs. While I-frames are sent frequently in video conferencing, our source image only needs to be sent once at the beginning. Moreover, the source image can be an image of the same person captured on a different day, a different person, or even a face portrait painting.

\vspace{1mm}
\noindent{\bf Adaptive number of keypoints.} Our basic model uses a fixed number of keypoints during training and inference. However, since the transmitted bits are proportional to the number of keypoints, it is advantageous to change this number to accommodate varying bandwidth requirements dynamically. Using keypoint dropouts at training time, we derive a model that can dynamically use a smaller number of keypoints for reconstruction. This allows us to transmit even fewer bits without compromising visual quality. On average, with the adaptive scheme, the number of sent keypoints is reduced from $K=20$ to $11.52$ (\supplementary{C.2}).

\vspace{1mm}
\noindent{\bf Benchmark dataset.} We manually select a dataset of $222$ high-quality talking-head videos. Each video's resolution is $512\times512$ and the length is up to $1024$ frames for evaluation.

\vspace{1mm}
\noindent{\bf Baselines.}
We compare our video streaming method with the popular H.264 codec. In order to conform to real-time video streaming cases, we disable the use of bidirectional B-frames, as this uses information from the future. By varying the constant rate factor (CRF) while encoding the ground truth input videos, we can obtain a set of videos of varying qualities and sizes suitable for a range of bandwidth availability.
We also compare with FOMM~\cite{siarohin2019first} and fs-vid2vid~\cite{wang2019few}, which also use keypoints or facial landmarks. For a fair comparison, we also compress their keypoints and Jacobians using our entropy coding scheme.

\vspace{1mm}
\noindent{\bf Metrics.}
We compare the compression effectiveness using the average number of bits required per pixel (bpp) for each output frame. We measure the compression quality using both automatic and human evaluations. Unlike traditional compression methods, our method does not reproduce the input image in a pixel-aligned manner but can faithfully reproduce facial motions and gestures. Metrics based on exact pixel alignments are ill-suited for measuring the quality of our output videos. We hence use the LPIPS perceptual similarity metric~\cite{zhang2018unreasonable} for measuring compression quality~\cite{blau2019rethinking}.

As shown on the left side of Fig.~\ref{fig:compression_metrics}, compared to the other neural talking-head synthesis methods, ours obtains better quality while requiring much lower bandwidth. This is because other methods send the full keypoints~\cite{siarohin2019first,wang2019few} and Jacobians~\cite{siarohin2019first}, while ours only sends the head pose and keypoint deformations. Compared with H.264 videos of the same quality, ours requires significantly lower bandwidth. For human evaluation, we show MTurk workers two videos side-by-side, one produced by H.264 and the other produced by our method's adaptive version. We then ask the workers to choose the video that they feel is of better quality. The preference scores are visualized on the right side of Fig.~\ref{fig:compression_metrics}.
Based on these scores, our compression method is comparable to the H.264 codec at a CRF value of $36$, which means our adaptive and $20$ keypoint scheme obtains $10.37\times$ and $6.5\times$ reduction in bandwidth compared to the H.264 codec, respectively.
To handle challenging corner cases for our video conferencing system and out-of-distribution videos, we further develop a binary latent encoding network that can efficiently encode the residual at the expense of additional bandwidth, the details of which are in \supplementary{C.3}.

\vspace{-.05in}
\section{Conclusion}

In this work, we present a novel framework for neural talking-head video synthesis and compression. We show that by using our unsupervised 3D keypoints, we are able to decompose the representation into person-specific canonical keypoints and motion-related transformations. This decomposition has several benefits: By modifying the keypoint transformation only, we are able to generate free-view videos. By transmitting just the keypoint transformations, we can achieve much better compression ratios than existing methods. These features provide users a great tool for streaming live videos. By dramatically reducing the bandwidth and ensuring a more immersive experience, we believe this is an important step towards the future of video conferencing.

\clearpage
\ifdefined\arxiv
\noindent{\bf Acknowledgements.}
We thank Jan Kautz for his valuable comments throughout the development of the work. We thank {Timo Aila}, {Koki Nagano}, {Sameh Khamis}, {Jaewoo Seo}, and {Xihui Liu} for providing very useful feedback to shape our draft. We thank {Henry Lin}, {Rochelle Pereira}, {Santanu Dutta}, {Simon Yuan}, {Brad Nemire},  {Margaret Albrecht}, {Siddharth Sharma}, and {Eric Ladenburg} for their~helpful~suggestions~in~presenting~our~visualization results.

\fi
{\small
\bibliographystyle{ieee_fullname}
\bibliography{gan}
}

\clearpage
\ifdefined\arxiv

\appendix
\section{Additional Network and Training Details}
\label{sec:network_details}

Here, we present the architecture of our neural talking-head model. We also discuss the training details. 

\subsection{Network architectures}

The implementation details of the networks in our model are shown in Fig.~\ref{fig:arch} and described below.

\vspace{1mm}
\noindent{\bf Appearance feature extractor $F$.}
The network extracts 3D appearance features from the source image. It consists of a number of downsampling blocks, followed by a convolution layer that projects the input 2D features to 3D features. We then apply a number of 3D residual blocks to compute the final 3D features $f_{s}$.

\vspace{1mm}
\noindent{\bf Canonical keypoint detector $L$.} The network takes the source image and applies a U-Net style encoder-decoder to extract canonical keypoints. Since we need to extract 3D keypoints, we project the encoded features to 3D through a $1\times 1$ convolution. The output of the $1\times 1$ convolution is the bottleneck of the U-Net. The decoder part of the U-Net consists of 3D convolution and upsampling layers.

\vspace{1mm}
\noindent{\bf Head pose estimator $H$ and expression deformation estimator $\Delta$.} We adopt the same architecture as in Ruiz~\etal~\cite{ruiz2018fine}. It consists of a series of ResNet bottleneck blocks, followed by a global pooling to remove the spatial dimension. Different linear layers are then used to estimate the rotation angles, the translation vector, and the expression deformations. The full angle range is divided into $66$ bins for rotation angles, and the network predicts which bin the target angle is in. The estimated head pose and deformations are used to transform the canonical keypoints to obtain the source or driving keypoints.

\vspace{1mm}
\noindent{\bf Motion field estimator $M$.}
After the keypoints are predicted, they are used to estimate warping flow maps. We generate a warping flow map $w_k$ based on the $k$-th keypoint using the first-order approximation~\cite{siarohin2019first}. Let $p_d$ be a 3D coordinate in the feature volume of the driving image $d$. The $k$-th flow field maps $p_d$ to a 3D coordinate in the 3D feature volume of the source image $s$, denoted by $p_s$, by:
\begin{equation}
w_k: R_{s} R_{d}^{-1} (p_d - x_{d,k}) + x_{s,k} \mapsto p_s.
\end{equation}
This builds a correspondence between the source and driving.

Using the flow field $w_k$ obtained from the $k$-th keypoint pair, we can warp the source feature $f_{s}$ to construct a candidate warped volume, $w_k(f_{s})$. After we obtain the warped source features $w_k(f_s)$ using all $K$ flows, they are concatenated together and fed to a 3D U-Net to extract features. Then a softmax function is employed to obtain the flow composition mask $m$, which consists of $K$ 3D masks, $\{m_1, m_2, ...,m_K\}$. These maps satisfy the constraints that $\sum_{k} m_k(p_d) = 1$ and $0\le m_k(p_d)\le 1$ for all $p_d$. These $K$ masks are then linearly combined with the $K$ warping flow maps, $w_k$'s, to construct the final warping map $w$ by $\sum_{k=1}^{K} m_k(p_d) w_k(p_d)$. To handle occlusions caused by the warping, we also predict a 2D occlusion mask $o$, which will be inputted to the generator $G$.

\vspace{1mm}
\noindent{\bf Generator $G$.} The generator takes the warped 3D appearance features $w(f_s)$ and projects them back to 2D. Then, the features are multiplied with the occlusion mask $o$ obtained from the motion field estimator $M$. Finally, we apply a series of 2D residual blocks and upsamplings layers to obtain the final image.

\begin{figure*}[t!]
    \centering
    \includegraphics[width=\textwidth]{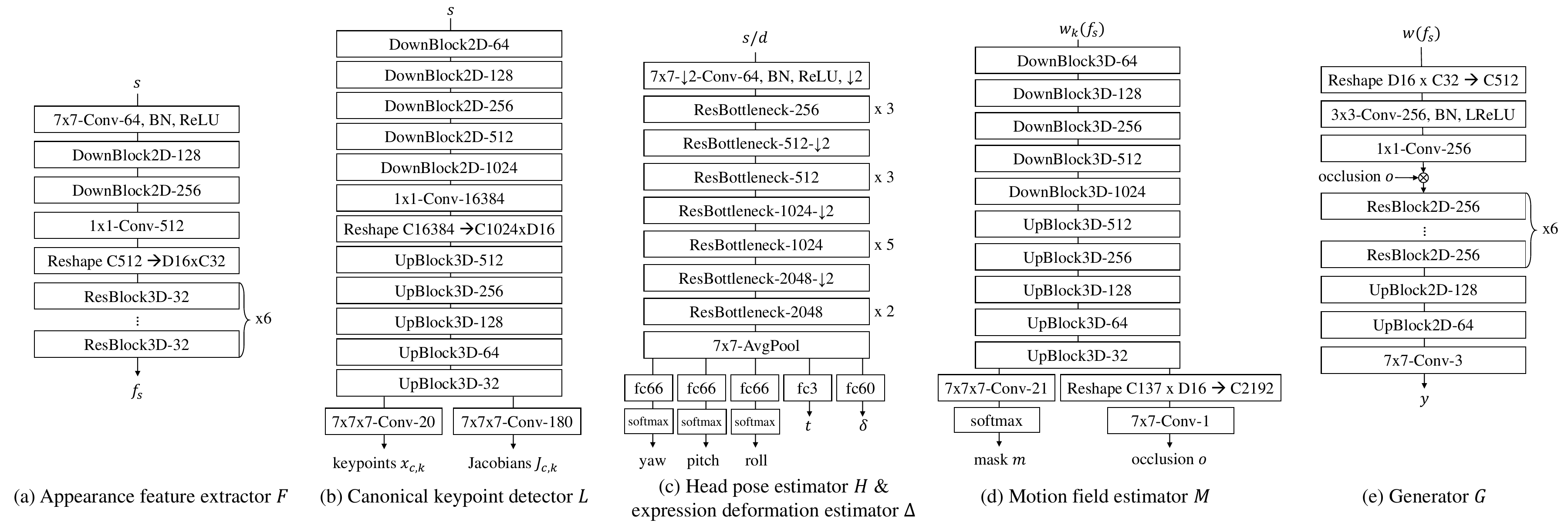}
    \vspace{-.1in}
    \caption{Architectures of individual components in our model. For the building blocks, please refer to Fig.~\ref{fig:arch_base}}
    \label{fig:arch}
    \vspace{-.1in}
\end{figure*}

\begin{figure}[t!]
    \centering
    \includegraphics[width=.4\textwidth]{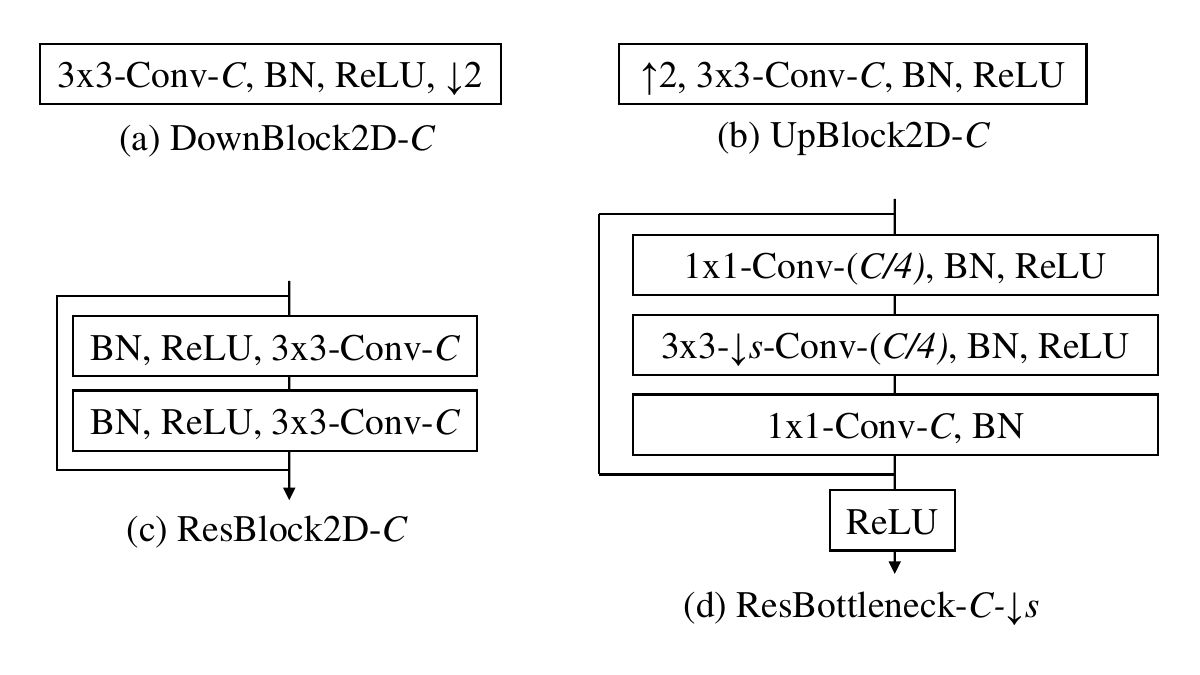}
    \vspace{-.1in}
    \caption{Building blocks of our model. For the 3D counterparts, we simply replace 2D convolutions with 3D convolutions in the blocks.}
    \label{fig:arch_base}
    \vspace{-.1in}
\end{figure}

\subsection{Losses}

We present details of the loss terms in the following.

\vspace{1mm}
\noindent{\bf Perceptual loss $\mathcal{L}_P$.} We use the multi-scale implementation introduced by Siarohin~\etal~\cite{siarohin2019first}. 
In particular, a pre-trained VGG network is used to extract features from both the ground truth and the output image, and the $L_1$ distance between the features is computed. Then both images are downsampled, and the same VGG network is used to extract features and compute the $L_1$ distance again. This process is repeated $3$ times to compute losses at multiple image resolutions. We use layers \texttt{relu\_1\_1, relu\_2\_1, relu\_3\_1, relu\_4\_1, relu\_5\_1} of the VGG19 network with weights $0.03125, 0.0625, 0.125, 0.25, 1.0$, respectively.
Moreover, since we are synthesizing face images, we also compute a single-scale perceptual loss using a pre-trained face VGG network~\cite{parkhi2015deep}. These losses are then summed together to give the final perceptual loss.

\vspace{1mm}
\noindent{\bf GAN loss $\mathcal{L}_{G}$.} We adopt the same patch GAN implementation as in~\cite{wang2018high,park2019semantic}, and use the hinge loss. Feature matching~\cite{wang2018high} loss is also adopted to stabilize training. We use single-scale discriminators for training $256\times 256$ images, and two-scale discriminators~\cite{wang2018high} for $512\times 512$ images.

\vspace{1mm}
\noindent{\bf Equivariance loss $\mathcal{L}_E$.} This loss ensures the consistency of estimated keypoints~\cite{zhang2018unsupervised,siarohin2019first}. In particular, let the original image be $d$ and its detected keypoints be $x_d$. When a known spatial transformation $\mathbf{T}$ is applied on image $d$, the detected keypoints $x_{\mathbf{T}(d)}$ on this transformed image $\mathbf{T}(d)$ should be transformed in the same way. Based on this observation, we minimize the $L_1$ distance $\|x_d -  \mathbf{T}^{-1}(x_{\mathbf{T}(d)})\|_1$. 
Affine transformations and randomly sampled thin plate splines are used to perform the transformation. 
Since all these are 2D transformations, we project our 3D keypoints to 2D by simply dropping the $z$ values before computing the losses.

\vspace{1mm}
\noindent{\bf Keypoint prior loss $\mathcal{L}_L$.} 
As described in the main paper, we penalize the keypoints if the distance between any pair of them is below some threshold $D_t$, or if the mean depth value deviates from a preset target value $z_t$.
In other words,
\begin{align}
    \mathcal{L}_L = \sum_{i=1}^K\sum_{j=1}^K \max(0, D_t - \|{x}_{d,i} - {x}_{d,j}\|_2^2)
+ \|Z({x}_{d}) - z_t\|
\end{align}
where $Z(\cdot)$ extracts the mean depth value of the keypoints. This ensures the keypoints are more spread out and used more effectively. We set $D_t$ to $0.1$ and $z_t$ to $0.33$ in our experiments.

\vspace{1mm}
\noindent{\bf Head pose loss $\mathcal{L}_H$.} 
We compute the $L_1$ distance between the estimated head pose $R_d$ and the one predicted by a pre-trained pose estimator $\bar{R}_d$, which we treat as ground truth. In other words,
$\mathcal{L}_H = \|R_d - \bar{R}_d\|_1$, where the distance is computed as the sum of differences of the Euler angles.

\vspace{1mm}
\noindent{\bf Deformation prior loss $\mathcal{L}_{\Delta}$.} 
Since the expression deformation $\Delta$ is the deviation from the canonical keypoints, their magnitude should not be too large. To ensure this, we put a loss on their $\mathcal{L}_1$ norm: $\mathcal{L}_{\Delta} = \|\delta_{d,k}\|_1$.

\vspace{1mm}
\noindent{\bf Final loss}
The final loss is given by:
\begin{align}
\mathcal{L} = &\lambda_P\mathcal{L}_{P}(d, y) + \lambda_G\mathcal{L}_{G}(d, y) +
\lambda_E\mathcal{L}_{E}(\{x_{d,k}\}) +
\nonumber\\
&\lambda_L\mathcal{L}_{L}(\{ x_{d,k}\}) +
\lambda_H\mathcal{L}_{H}(R_d, \bar{R}_d) +
\lambda_{\Delta}\mathcal{L}_{\Delta}(\{\delta_{d,k}\})
\end{align}
where $\lambda$'s are the weights and are set to $10, 1, 20, 10, 20, 5$ respectively in our implementation.

\subsection{Optimization}

We adopt the ADAM optimizer~\cite{kingma2014adam} with $\beta_1=0.5$ and $\beta_2=0.999$. The learning rate is set to $0.0002$. We apply Spectral Norm~\cite{miyato2018spectral} to all the layers in both the generator and the discriminator. We use synchronized BatchNorm for the generator. Training is conducted on an NVIDIA DGX1 with 8 32GB V100 GPUs.

We adopt a coarse-to-fine approach for training. We first train our model on $256\times 256$ images for $100$ epochs. We then finetune on $512\times 512$ images for another $10$ epochs.
\section{Additional Experiment Details}

\subsection{Datasets}

We use the following datasets in our evaluations.

\vspace{1mm}
\noindent{\bf VoxCeleb2~\cite{chung2018voxceleb2}.} The dataset contains about 1M talking-head videos of different celebrities. We follow the training and test split proposed in the original paper. where we use 280K videos with high bit-rates to train our model. We report our results on the validation set, which contains about 36k videos.

\vspace{1mm}
\noindent{\bf TalkingHead-1KH.}
We compose a dataset containing about $1000$ hours of videos from various sources. A large portion of them is from the YouTube website with the creative common license. We also use videos from
the Ryerson audio-visual dataset~\cite{livingstone2018ryerson} as well as a set of videos that we recorded with the permission from the subject ourselves. We only use videos whose resolution and bit-rate are both high. We call this dataset \emph{TalkingHead-1KH}. The videos in the TalkingHead-1KH are in general with higher resolutions and better image quality than those in the VoxCeleb2.

\subsection{Metrics}

We use a set of metrics to evaluate a talking-head synthesis method. We use $L_1$, PSNR, SSIM, and MS-SSIM for quantifying the faithfulness of the recreated videos. We use FID to measure how close is the distribution of the recreated videos to that of the original videos. We use AKD to measure how close the facial landmarks extracted by an off-the-shelf landmark detector from the recreated video are to those in the original video. In the following, we discuss the implementation details of these metrics.

\vspace{1mm}
\noindent{\bf $L_1$.} We compute the average $L_1$ distance between generated and real images.

\vspace{1mm}
\noindent{\bf PSNR} measures the image reconstruction quality by computing the mean squared error (MSE) between the ground truth and the reconstructed image.

\vspace{1mm}
\noindent{\bf SSIM/MS-SSIM.} SSIM measures the structural similarity between patches of the input images. Therefore, it is more robust to global illumination changes than PSNR, which is based absolute errors. MS-SSIM is a multi-scale variant of SSIM that works on multiple scales of the images and has been shown to correlate better with human perception.

\vspace{1mm}
\noindent{\bf FID~\cite{heusel2017gans}} measures the distance between the distributions of synthesized and real images. We use the pre-trained InceptionV3 network to extract features from both sets of images and estimate the distance between them.

\vspace{1mm}
\noindent{\bf Average keypoint distance (AKD).} We use a facial landmark detector~\cite{dlib2009} to detect landmarks of real and synthesized images and then compute the average distance between the corresponding landmarks in these two images.

\subsection{Relative motion transfer}
For cross-identity motion transfer results in our experiment section, we transfer absolute motions in the driving video. For completeness, we also report quantitative comparisons using relative motion proposed in~\cite{siarohin2019first} in Table~\ref{tbl::relative_cross_id}. As can be seen, our method still performs the best.

\begin{table}[tb!]
    \caption{Cross-identity transfer using relative motion.}
    \label{tbl::relative_cross_id}
	\centering
	\small
	\resizebox{.8\columnwidth}{!}{%
    \begin{tabular}{rcccc}
    \toprule
    & \multicolumn{2}{c}{VoxCeleb2} & \multicolumn{2}{c}{TalkingHead-1KH} \\
    Method & FID$\downarrow$ & CSIM$\uparrow$ & FID$\downarrow$ & CSIM$\uparrow$ \\
    \cmidrule(r){2-3} \cmidrule(r){4-5}
    fs-vid2vid~\cite{wang2019few} & 48.48 & 0.928 & 44.83 & 0.955 \\
    FOMM~\cite{siarohin2019first} & 48.91 & 0.954 & 42.26 & 0.961 \\
    {\bf Ours } & {\bf 46.43} & {\bf 0.960} & {\bf 41.25} & {\bf 0.964} \\
    \bottomrule
	\end{tabular}
	}
	\vspace{-.1in}
\end{table}

\subsection{Ablation study}
\begin{table}[tb!]
    \vspace{-.1in}
	\caption{Ablation study. Compared with all the other alternatives, our model (the preferred setting) works the best.}
\label{tbl::ablation}
	\centering
	\small
	\tabcolsep=0.11cm
    \begin{tabular}{rcccccc}
\toprule
Method & L1 & PSNR & SSIM & MS-SSIM & FID & AKD \\
\cline{1-7}
Direct pred. &
10.84 & 24.00 & 0.80 & 0.83 & 58.55 & 4.26 \\
{\bf Ours (20 kp)} & 
{\bf 10.67} & {\bf 24.20} & {\bf 0.81} & {\bf 0.84} & {\bf 52.08} & {\bf 3.74} \\
\midrule
2D Warp &
11.64 & 23.38 & 0.79 & 0.82 & 58.75 & 4.20 \\
{\bf Ours (20 kp)} & 
{\bf 10.67} & {\bf 24.20} & {\bf 0.81} & {\bf 0.84} & {\bf 52.08} & {\bf 3.74} \\
\midrule
10 kp &
11.49 & 23.36 & 0.79 & 0.82 & 56.27 & 4.31 \\
15 kp &
11.35 & 23.53 & 0.79 & 0.82 & 54.36 & 4.50 \\
{\bf Ours (20 kp)} & 
{\bf 10.67} & {\bf 24.20} & {\bf 0.81} & {\bf 0.84} & {\bf 52.08} & {\bf 3.74} \\
\bottomrule
	\end{tabular}
 \end{table}
We perform the following ablation studies to verify the effectiveness of our several important design choices.

\vspace{1mm}
\noindent{\bf Two-step vs.\ direct keypoint prediction.} We estimate the keypoints in an image by first predicting the canonical keypoints and then applying the transformation and the deformations. To compare this approach with direct keypoint location prediction, we train another network that directly predicts the final source and driving keypoints in the image. In particular, the keypoint detector $L$ directly predicts the final source and driving keypoints in the image instead of the canonical ones, and there is no pose estimator $H$ and deformation estimator $\Delta$. Since there is no pose estimator, we do not need any pose supervision (i.e., head pose loss) for this direct prediction network. Note that while this model is only slightly inferior to our final (two-step) model quantitatively, it has no pose control for the output video since the head pose is no longer estimated, so a major feature of our method would be lost.

\vspace{1mm}
\noindent{\bf 3D vs.\ 2D warping.} We generate 3D flow fields from the estimated keypoints to warp 3D features. Another option is to project the keypoints to 2D, estimate a 2D flow field, and extract 2D features from the source image. The estimated 2D flow filed is then used to warp 2D image features.

\vspace{1mm}
\noindent{\bf Number of keypoints.} We show that our approach's output quality is positively correlated with the number of keypoints.

As can be seen in Table~\ref{tbl::ablation}, our model works better than all the other alternatives on all of the performance metrics.

\subsection{Failure cases}

While our model is in general robust to different situations, it cannot handle large occlusions well. For example, when the face is occluded by the person's hands or other objects, the synthesis quality will degrade, as shown in Fig.~\ref{fig:failure}

\begin{figure}[t!]
    \vspace{-.1in}
    \centering
    \includegraphics[width=.45\textwidth]{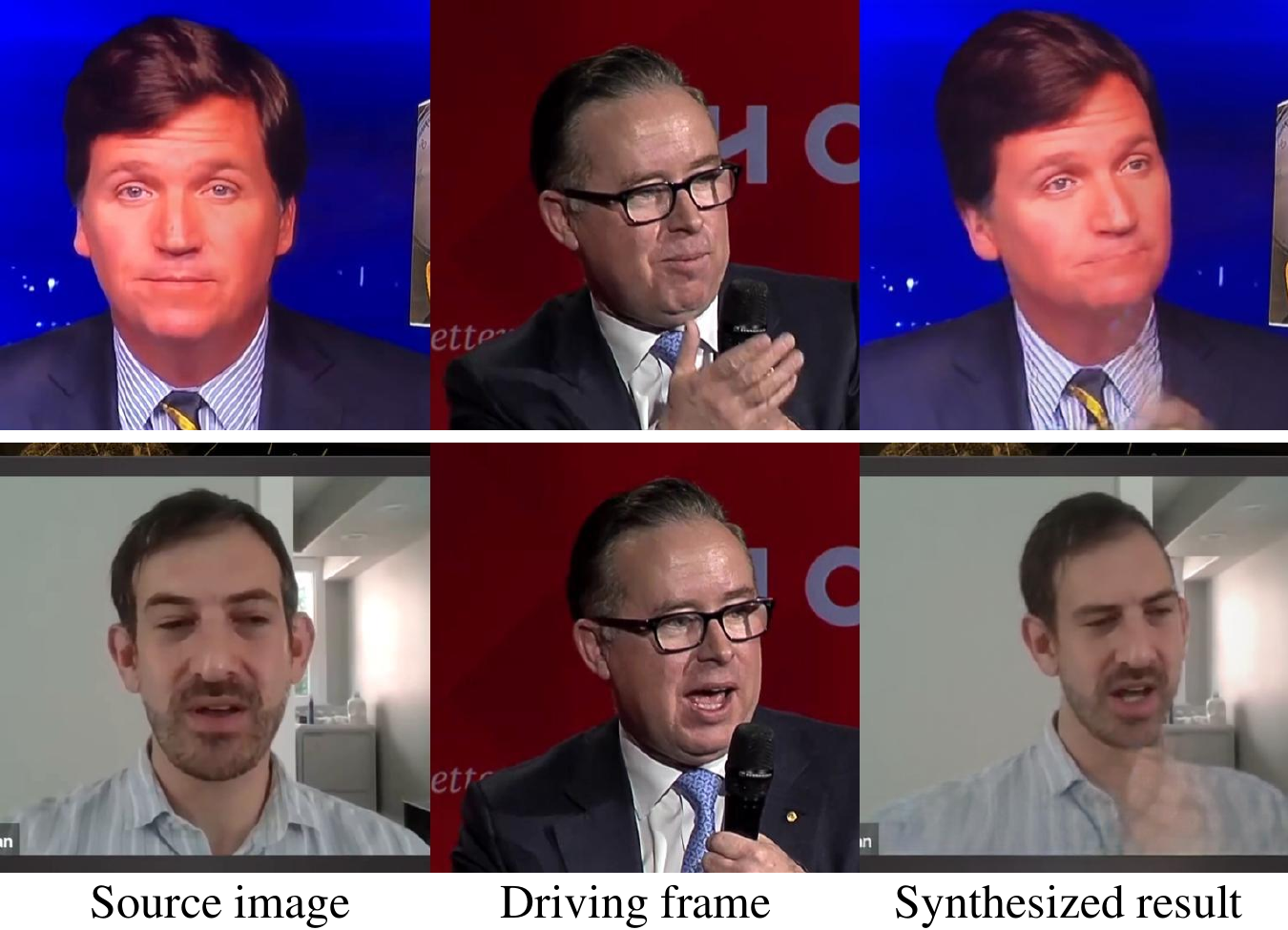}
    \vspace{-.1in}
    \caption{Example failure cases. Our method still struggles when there are occluders such as hands in the image.}
    \label{fig:failure}
    \vspace{-.1in}
\end{figure}

\subsection{Canonical keypoints for face recognition}

Our canonical keypoints are formulated to be independent of the pose and expression change. They should only contain a person's geometry signature, such as the shapes of face, nose, and eyes. To verify this, we conduct an experiment using the canonical keypoints for face recognition.

We extract canonical keypoints from $384$ identities in the VoxCeleb2~\cite{chung2018voxceleb2} dataset to form a training set. For each identity, we also pick a different video of the same identity to form the test set. The training and test videos of the same subject have different head poses and expressions. A face recognition algorithm would fail if it could not filter out pose and expression information. To prove our canonical keypoints are independent to poses and expressions, we apply a simple nearest neighbor classifier using our canonical keypoints for the face recognition task.

Overall, our canonical keypoints reaches an accuracy of $0.070$, while a random guess has an accuracy of $0.0026$ (Ours is $27\times$ better than the random guess.). On the other hand, a classifier using the off-the-shelf dlib landmark detector only achieves an accuracy of $0.013$, which means our keypoints are $5\times$ more effective for face recognition. 

\section{Additional Video Conferencing Details}

\begin{table}[t]
    \caption{Size of per-frame metadata in bytes for talking-head methods before and after arithmetic compression.}
	\label{table:arithmetic_compression}	
    \centering
    \resizebox{\columnwidth}{!}{%
    \begin{tabular}{rccccc}
        \toprule
        \multirow{2}{*}{Method} & Before & \multicolumn{4}{c}{After Compression} \\
        & Compression & Mean & Min & Max & Median \\
        \midrule
        fs-vid2vid~\cite{wang2019few} & 504 & 231.42 & 158 & 599 & 238 \\
        FOMM~\cite{siarohin2019first} & 240 & 171.09 & 159 & 210 & 169 \\
        Ours (20 kp) & 132 & 84.44 & 78 & 104 & 84 \\
        Ours (adaptive) & 81.16 & 53.03 & 25 & 102 & 45 \\
        \bottomrule
    \end{tabular}%
    }
\end{table}
\begin{figure*}[!t]
	\vspace{-.1in}
	\centering
	\setlength{\tabcolsep}{0pt}

	\begin{tabular}{C{0.16\textwidth}|C{0.16\textwidth}|C{0.16\textwidth}|C{0.16\textwidth}|C{0.16\textwidth}|C{0.16\textwidth}}
		\multirow{2}{*}{Ground truth} & \multirow{2}{*}{Compressed} & Compressed + Decoded residual &
		\multirow{2}{*}{Ground truth} & \multirow{2}{*}{Compressed} & Compressed + Decoded residual \\
		\multicolumn{3}{c}{\includegraphics[width=0.48\textwidth]{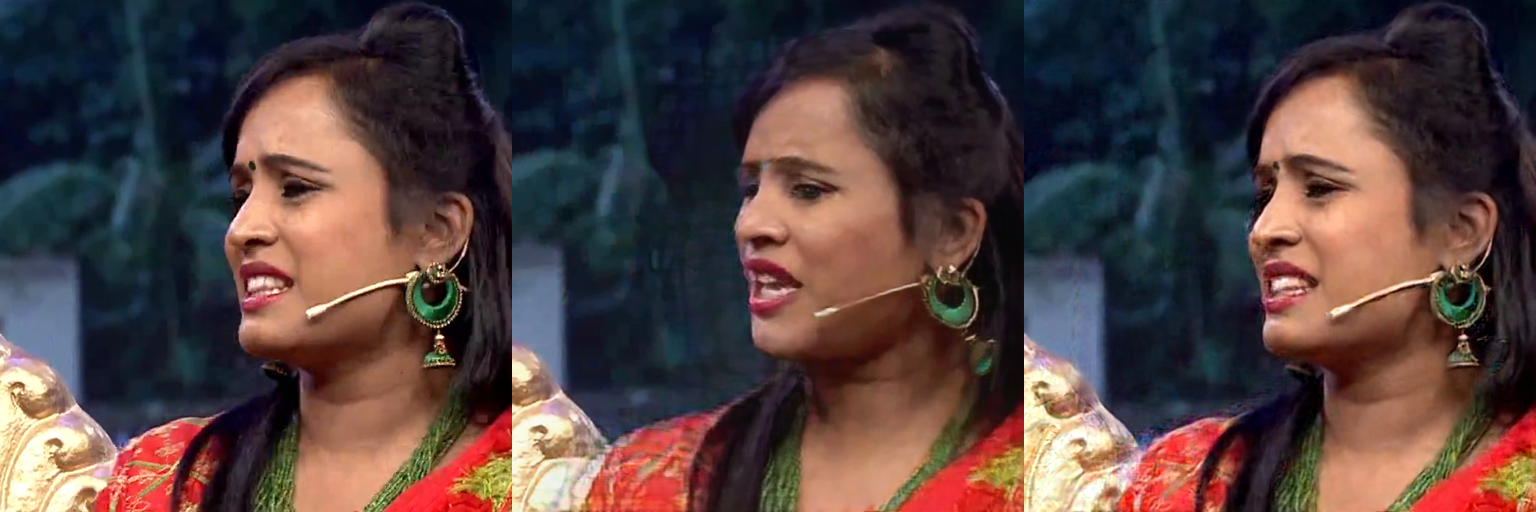}} & \multicolumn{3}{c}{\includegraphics[width=0.48\textwidth]{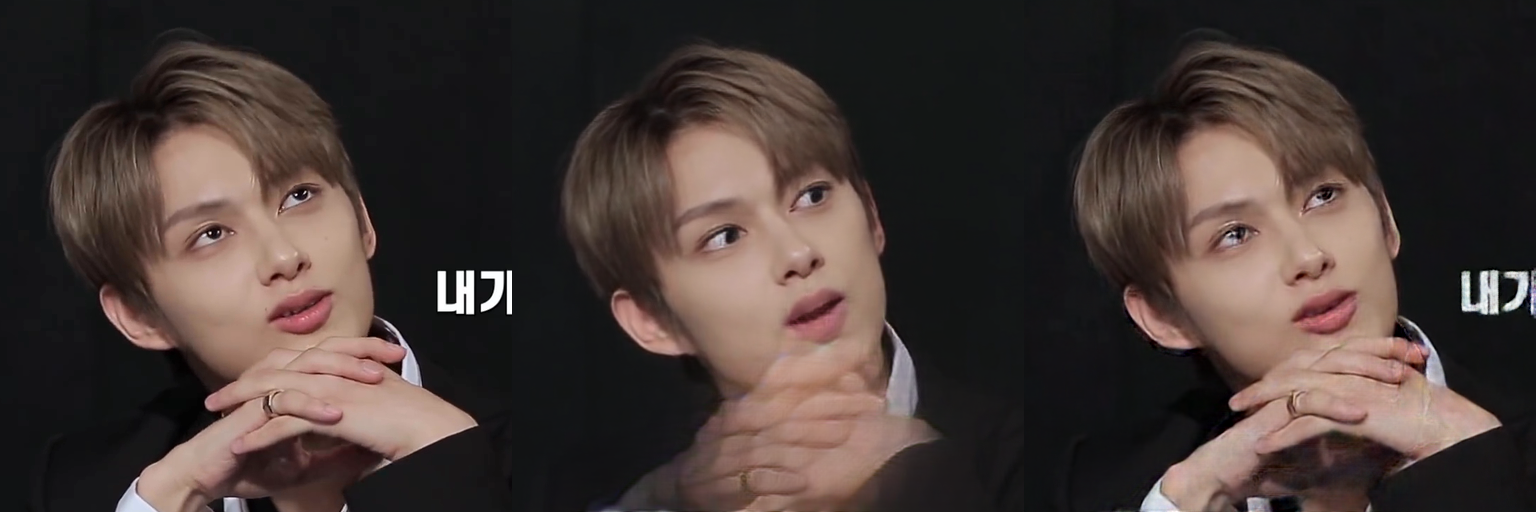}} \\
		\multicolumn{3}{c}{\includegraphics[width=0.48\textwidth]{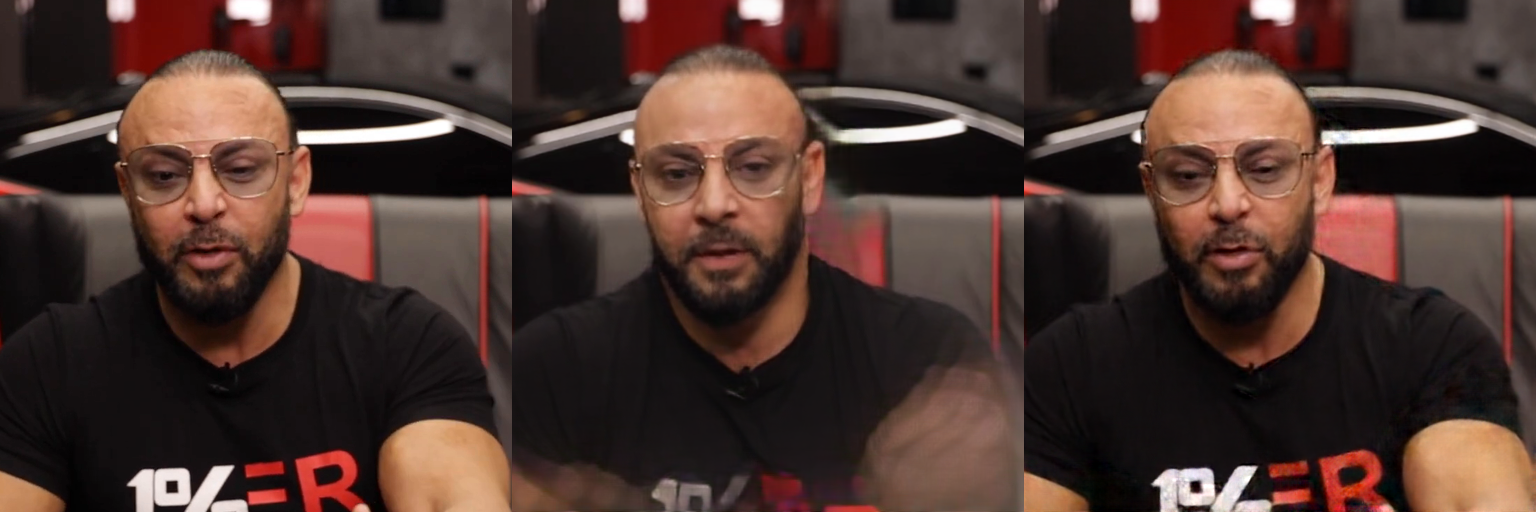}} & \multicolumn{3}{c}{\includegraphics[width=0.48\textwidth]{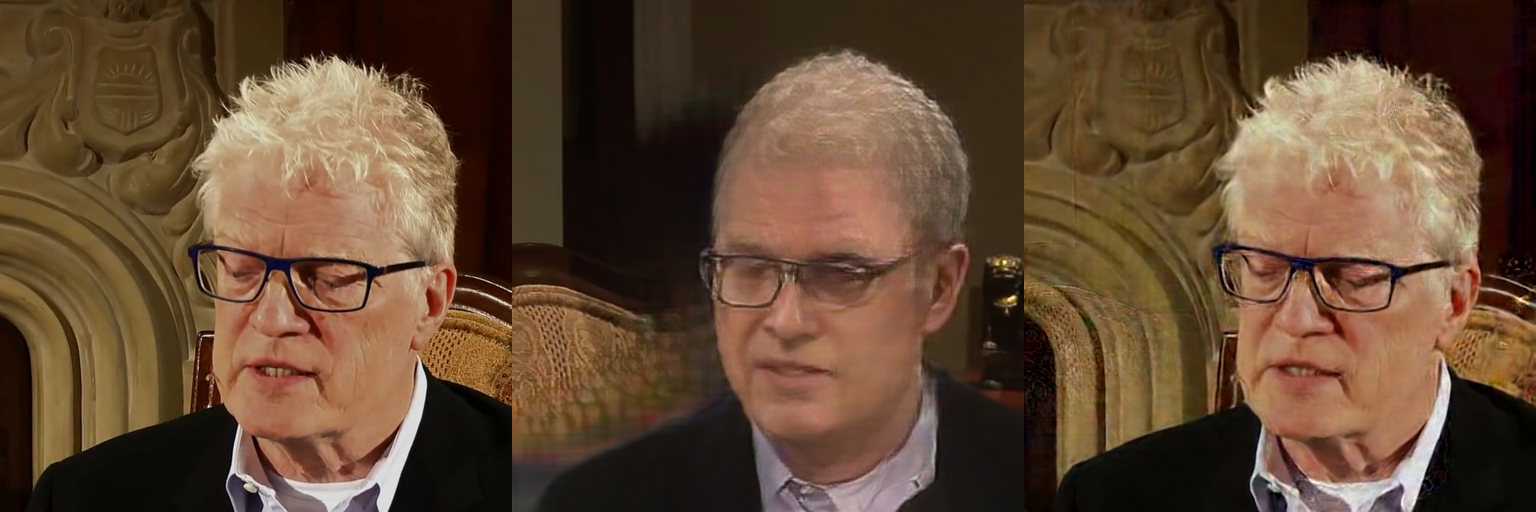}} \\
	\end{tabular}
	\vspace{-.1in}
	\caption{Fixing artifacts in compressed images using our binary residual encoder. We are able to fix artifacts caused due to the introduction of new objects, changes in background, as well as extreme poses by transmitting the residuals encoded as binary values. Each residual binary latent code requires only about 13.40 KB and can replace sending new source images.}
	\label{fig:residual}

	\vspace{10pt}

    \setlength{\tabcolsep}{0pt}
    \begin{tabular}{C{0.35\textwidth}C{0.3\textwidth}C{0.28\textwidth}}
        \multicolumn{3}{c}{\includegraphics[width=\textwidth]{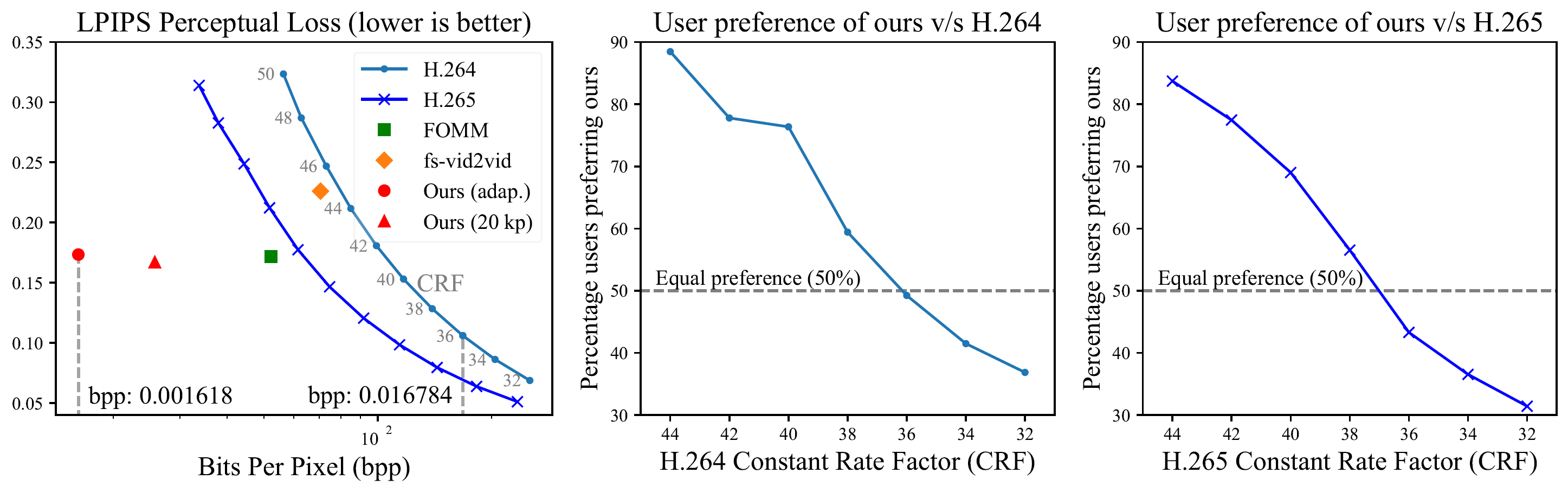}} \\[-5pt]
        (a) & (b) & (c) \\
    \end{tabular}
	\vspace{-.1in}
	\caption{Automatic and human evaluations for video compression. Ours requires much lower bandwidth than the H.264 and H.265 codecs and other talking-head synthesis methods thanks to our keypoint decomposition and adaptive scheme.}
	\label{fig:compression_metrics}
	\vspace{-.2in}
\end{figure*}

\subsection{Entropy encoder}
We represent each rotation angle, translation, and deformation value as an fp16 floating-point number. Each number consumes two bytes. Naively transmitting the $3K+6$ floating numbers will result in transmitting $6K+12$ bytes. We adopt arithmetic coding~\cite{langdon1984introduction} to encode the $3K+6$ numbers. Arithmetic coding is one kind of entropy coding. It assigns different codeword lengths to different symbols based on their frequencies. The symbol that appears more often will have a shorter code.

We first apply the driving image encoder to a validation set of $127$ videos that are not included in the test set. Each frame will give us $6K+12$ bytes. We treat each of the bytes separately and build a frequency table for each byte. This gives us $6K+12$ frequency tables.
When encoding the test set, we encode each byte using the associated frequency table learned from the validation set. This results in a varying-length representation that is much smaller than $6K+12$ bytes on average.

Table~\ref{table:arithmetic_compression} shows the sizes of the per-frame metadata in bytes that needs to be transmitted for various talking-head methods before and after performing the arithmetic compression for an image size of 512$\times$512. Our adaptive scheme requires a per-frame metadata size of {$53.03$ B}, which corresponds to $(53.03 \times 8 / 512^2) = 0.001618$ bits per pixel.

\subsection{Adaptive number of keypoints}
Our basic model uses a fixed number of keypoints during training and inference. However, on a video call, it is advantageous to adaptively change the number of keypoints used to accommodate varying bandwidth and internet connectivity. We devise a scheme where our synthesis model can dynamically use a smaller number of keypoints for reconstruction. This is based on the intuition that not all of the images are of the same complexity. Some just require fewer keypoints. Using fewer keypoints, we can reduce the bandwidth required for video conferencing because we just need to send a subset of $\delta_{d,k}$'s. To train a model that supports a varying number of keypoints, we randomly choose an index into the array of ordered keypoints, and dropout all values from that index till the end of the array. This dropout percentage ranges from 0\% to 75\%. This scheme is also helpful when the available bandwidth suddenly drops.

\subsection{Binary encoding of the residuals}
When the contents of the video being streamed change drastically, \eg when new objects are introduced into the video or the person changes, it becomes necessary to update the source frame being used to perform the talking-head synthesis. This can be done by sending a new image to the receiver. We also devise a more efficient scheme to encode and send only the residual between the ground truth frame and the reconstructed frame, instead of an entirely new source image. To encode a residual image of size $512\times 512$, we use a 3-layer network with convolutions of kernel size 3, stride 2, and 32 channels, similar to the network proposed by Tsai~\etal~\cite{tsai2017learning}. We compute the sign of the latent code of size $32\times 64 \times 64$ to obtain binary latent codes. The decoder also consists of 3 convolutional layers of 128 filters and uses the pixel shuffle layer to perform upsampling. After arithmetic coding, the binary latent code requires 13.40 KB on average. Note that we do not need to send the encoded binary residual every frame. We just need to send it when the current source image is not good enough to reconstruct the current driving image. In the receiver side, we will use the encoded residual to improve the quality of the reconstructed image. The reconstructed image will become the new source image for decoding future frames using the encoded rotation, translation, and deformations. Example improvements after adding the residual are shown in Fig.~\ref{fig:residual}.

\subsection{Dataset}
\vspace{-.05in}

For testing, we collect a set of high-resolution talking-head videos from the web. We ensure that the head is of size at least 512$\times$512 pixels and manually check each video to ensure its quality. This results in a total of 222 videos, with a mean of 608 frames, a median of 661 frames, and a min and max of 20 and 1024 frames, respectively.

\vspace{-.05in}
\subsection{Additional experiment results}
\vspace{-.05in}

In Fig.~\ref{fig:compression_metrics}(a), we show the achieved LPIPS score by our approach under the adaptive setting (red circle), our approach under the 20 keypoint setting (red triangle), FOMM (green square), fs-vid2vid (orange diamond), H.264, and H.265 using different bpp rates. We observe that our method requires much lower bandwidth than the competing methods.

\vspace{1mm}
\noindent{\bf User study.} Here, we describe the details of our user study. We use the Amazon Mechanical Turk (MTurk) platform for the user preference score. A worker needs to have a lift-time approval rate greater than 98 to be qualified for our study. This means that the requesters approve 98\% of his/her task assignments. For comparing two competing methods, we generate 222 videos from each method. We show the corresponding pair of videos from two competing methods to three different MTurk workers and ask them to select which one has better visual quality. This gives 666 preference scores for each comparison. We report the average preference score achieved by our method. We compare our adaptive approach to both H.264 and H.265. The user preference scores of our approach when compared to H.264 and H.265 are shown in Fig.~\ref{fig:compression_metrics}(b) and (c), respectively. We found that our approach renders comparable performance to H.264 with CRF value 36. For H.265, our approach is comparable to CRF value 37. Our approach was able to achieve the same visual quality using a much lower bit-rate.

\fi
\end{document}